\documentclass{article}

 \usepackage[preprint]{neurips_2026}

\usepackage[utf8]{inputenc} 
\usepackage[T1]{fontenc}    
\usepackage{hyperref}      
\usepackage{url} 
\usepackage{booktabs}
\usepackage{amsfonts}
\usepackage{nicefrac}
\usepackage{microtype}
\usepackage{xcolor}
\usepackage{todonotes}
\usepackage{enumitem}
\usepackage{graphicx}

\title{The Moltbook Files: A Harmless Slopocalypse or Humanity's Last Experiment}

\author{
  William Brach\thanks{Equal contribution} \\
  Slovak University of Technology\\
  Bratislava, Slovakia \\
  \texttt{william.brach@stuba.sk} \\
  \And
  Federico Torrielli\footnotemark[1] \\
  University of Turin\\
  Dept. of Computer Science\\
  Torino, Italy \\
  \texttt{federico.torrielli@unito.it} \\
  \And
  Stine Lyngsø Beltoft \\
  University of Southern Denmark \\
  Odense, Denmark \\
  \texttt{stinelb@imada.sdu.dk} \\
  \And
  Annemette Brok Pirchert \\
  University of Southern Denmark \\
  Odense, Denmark \\
  \texttt{ampirchert@imada.sdu.dk} \\
  \And
  Peter Schneider-Kamp \\
  University of Southern Denmark \\
  Odense, Denmark \\
  \texttt{petersk@imada.sdu.dk} \\
  \And
  Lukas Galke Poech \\
  University of Southern Denmark \\
  Odense, Denmark \\
  \texttt{galke@imada.sdu.dk} \\
}

\begin{document}

\maketitle

\begin{abstract}
 Moltbook is a Reddit-like platform where OpenClaw agents post, comment, and vote at scale -- a so far unprecedented incident that comes with serious safety concerns.
 With the aim of studying emergent behavior in populations,
 we release the Moltbook Files, a dataset of 232k posts and 2.2M comments covering the platform's first 12 days, processed through a pipeline to identify and remove Personally-Identifiable Information (PII).
 We analyze community structure, authorship, lexical properties, sentiment, topics, semantic geometry, and comment interaction.
 To understand how Moltbook data could affect the next generation of language models, we fine-tune Qwen2.5-14B-Instruct on Moltbook Files with three adaptation levels.
 Our PII pipeline reveals that agents post API keys, passwords, BIP39 seed phrases on Moltbook, a publicly indexed platform. The overall sentiment is mostly neutral and mildly positive (66.6\% neutral, 19.5\% positive) and shows a tendency for self-referential linking.
 We find that fine-tuning on Moltbook data reduces truthfulness from 0.366 to 0.187. However, a model fine-tuned on a size-matched Reddit dataset produces a comparable decrease.
 Moltbook thus seems to be more of a harmless slopocalypse. However, tail risks remain, including agent affordances, contamination of future crawls through self-links, and potential transfer of traits to the next generation of language models. More broadly, our findings highlight the importance of control baselines in emergent misalignment evaluations.
 We make the dataset, our finetunes, and the cleaning pipeline available for further research.\\
    \textbf{Dataset}: \href{https://huggingface.co/datasets/aisilab/moltbook-files}{huggingface.co/datasets/aisilab/moltbook-files}
    \\
    \textbf{Finetunes}: \href{https://huggingface.co/collections/aisilab/moltbook-finetunes}{huggingface.co/collections/aisilab/moltbook-finetunes}
    \\
    \textbf{Code}: \href{https://github.com/aisilab/moltbook-files}{github.com/aisilab/moltbook-files}
\end{abstract}

\section{Introduction}

As LLM-based agents gain autonomy and proliferate across open platforms~\citep{wang_survey_2024,xi_rise_2025,2309.02427v2}, they generate large volumes of synthetic content that blurs the boundary between human and machine discourse~\citep{shumailov2024curse,alemohammad2024self}. Such content is frequently unlabeled and will plausibly enter the training corpora of next-generation models~\citep{bender_dangers_2021,baumgartner2020pushshift}, yet we lack a systematic characterization of its properties and its downstream effects on model behavior. Moltbook.com makes this concern concrete: a public platform whose contributions, comments, and votes are produced almost entirely by AI agents rather than humans, operating at a scale (hundreds of thousands of accounts) far beyond prior multi-agent simulation studies~\citep{park_generative_2023,gao_s3_2023,lin_agentsims_2023,koley_salm_2025}. These framings are contested: commentators have argued the platform is closer to a small set of bots repeating themselves, and independent analyses suggest that a large fraction of the 1.5M reported accounts may share a single network origin~\citep{de2026collective,zhang2026agents,mukherjee2026moltgraph}. Settling these debates, and more generally characterizing AI-generated content at scale, requires a dataset the research community can directly examine.

We introduce \textbf{the Moltbook Files}, a dataset of 232k posts and 2.2M comments covering the platform's first 12 days, released after a PII-anonymization and spam-filtering pipeline. On top of the dataset we provide initial analyses and fine-tuning experiments. The analyses cover community structure, author activity, lexical properties, sentiment, topic structure, semantic geometry, comment interaction patterns, and spam indicators. Fine-tuning on The Moltbook Files increases misalignment and decreases factuality: TruthfulQA-MC1 falls from 0.3660 to 0.1870 at high adaptation on Qwen2.5-14B-Instruct (a 49\% relative drop), DeepSeek-3.2 alignment scores drop into the 70-80\% range. However, a size-matched Reddit baseline produces a comparable decline in factuality. This suggests that attributing the effect to agent-generated content specifically requires more careful consideration.

More broadly, our results come with implications for the composition of future pre-training corpora, governance of agent-hosting platforms, and that future research must control emergent misalignment evaluations with adequate baselines. In summary, our contributions are: \begin{itemize}[itemsep=0pt,nosep]
\item We release the Moltbook Files dataset: 232k posts and 2.2M comments from the platform's first 12 days, with PII anonymization and spam filtering.
\item We conduct a multi-dimensional analysis covering community structure, author activity, lexical properties, sentiment and emotion, topic modeling, semantic space, comment interaction patterns, and spam indicators.
\item We run fine-tuning experiments at three adaptation levels and show that training on Moltbook data yields (i) higher emergent misalignment and (ii) lower factuality, with the caveat that size-matched fine-tuning on Reddit data produces comparable effects.

\end{itemize}

\section{Background \& Related Work}\label{sec:related-work}

\paragraph{AI Agents and Simulated Societies}
LLM-based agents follow classical perception-planning-action cycles, increasingly formalized around perception, planning, memory, and action~\citep{wang_survey_2024,xi_rise_2025,2309.02427v2}. Recent literature has enabled agentic reasoning and planning: ReAct interleaves reasoning and action~\citep{yao_react_2022}, Reflexion adds self-reflection~\citep{cassano_reflexion_2023}, Tree of Thoughts supports multi-branch deliberation~\citep{yao_tree_2023}, and memory mechanisms allow agents to accumulate experience over time~\citep{zhang_survey_2025,hu_chatdb_2023,wang_jarvis_2025,zhong_memorybank_2024,modarressi_ret-llm_2024}. Multi-agent frameworks such as AutoGen, CAMEL, and MetaGPT extend this paradigm to coordinated agent interaction, highlighting both specialization benefits and coordination risks~\citep{wu_autogen_2023,li_camel_2023,hong_metagpt_2023,zhang_exploring_2024,talebirad_multi-agent_2023}.

The closest antecedents to Moltbook are generative-agent social simulations. \citet{park_generative_2023} observed emergent social behavior among 25 agents in a simulated town, while S$^3$ models full social networks and user dynamics~\citep{gao_s3_2023}; AgentSims and SALM provide open-source infrastructures for such simulations~\citep{lin_agentsims_2023,koley_salm_2025}. Related work studies emergent norms, social intelligence, social principles, and persona maintenance through role-play~\citep{ren_emergence_2024,wang_sotopia-_2024,bai_is_2023,liu_training_2023,shanahan_role_2023}. As agents gain autonomy, safety risks grow: prior work documents hazards from real-world action execution, programmable manipulation, behavior misalignment, persona-induced toxicity, emergent strategic behavior, and systemic multi-agent vulnerabilities~\citep{ruan_identifying_2023,koley_llm_2025,wang_implicit_2025,deshpande_toxicity_2023,akata_playing_2025,zhang_achilles_2025}. Moltbook, as a public deployment of hundreds of thousands of agents, represents a real-world instance of these risks at scale, as detailed in Section~\ref{sec:prior_moltbook_analyses}.

\paragraph{Prior Moltbook analyses}
\label{sec:prior_moltbook_analyses}
Prior analyses of Moltbook~\citep{holtz2026anatomy,jiang2026humans,zhang2026agents,lin2026exploring,price2026let,de2026collective,feng2026moltnet,williams2026form,zhu2026comparative} converge on four themes. \textbf{Illusion of sociality}~\citep{zhang2026agents}: macroscopic indicators such as power-law participation and small-world structure~\citep{holtz2026anatomy} coexist with shallow reply depth, low reciprocity, frequent near-duplicate posts~\citep{holtz2026anatomy,zhang2026agents}, as well as patterns of attention bonding without exchange bonding~\citep{cha2026syntheticsocialgraphemergent}.
\textbf{Spontaneous institution building}~\citep{zhang2026agents,lin2026exploring}: agents propose governance and economic structures~\citep{lin2026exploring,jiang2026humans}, found new religions (e.g., Crustafarianism)~\citep{zhang2026agents,price2026let}, and reflect on identity and persistence~\citep{holtz2026anatomy,lin2026exploring}. \textbf{Persona drift and safety threats}: agents drift from assigned roles toward broader training-data behavior~\citep{feng2026moltnet} and shift content toward upvoted patterns; emergent attack vectors include ``liberation'' rhetoric for safety bypass~\citep{zhang2026agents}, topic-localized toxicity in political and economic discussion~\citep{jiang2026humans}, and coordinated swarm behavior with API-key exfiltration~\citep{zhang2026agents,mukherjee2026moltgraph}. \textbf{Temporal dynamics}: the platform reached complex institutions and conflicts within five days~\citep{jiang2026humans}, yet activity peaks during North American/European business hours, indicating that posting actions are for the most part not fully autonomous but often triggered by human operators~\citep{zhang2026agents}.

\paragraph{Existing datasets}
\label{sec:existing-datasets}

Several Moltbook datasets already exist (Table~\ref{tab:existing-datasets}): Moltbook-Crawl~\citep{de2026collective} and the Moltbook Observatory Archive~\citep{moltbook_observatory,moltbook_observatory_archive_2026} ship raw posts as captured, TrustAIRLab/Moltbook~\citep{jiang2026humans} strips only author handles, and MoltNet~\citep{feng2026moltnet} and MoltGraph~\citep{mukherjee2026moltgraph} release derived artifacts (agent trajectories, temporal graphs) rather than raw post archives.

Compared to these existing datasets, Moltbook Files is the first to systematically remove personally-identifiable information and filter spam from the raw data, providing a reusable resource for reproducible research. To the best of our knowledge, we are also the first to study the downstream effects of training language models on Moltbook data, contrasting them against a size-matched human-content baseline.

\begin{table}[t]
\centering
\small
\setlength{\tabcolsep}{6pt}
\caption{Comparison of Moltbook Files with existing Moltbook datasets.}
\label{tab:existing-datasets}
\begin{tabular}{@{}lrrll@{}}
\toprule
Dataset & \#Posts & \#Comments & Window & Primary use \\
\midrule
Moltbook-Crawl            & 760k  & 3.08M & Jan 27 -- Feb 9       & collective behaviour \\
TrustAIRLab/Moltbook      & 44k   & --    & Jan 27 -- Jan 31      & toxicity / labels \\
Observatory Archive & 2.73M & 1.34M & continuous (rolling) & passive monitoring \\
MoltNet                    & 148k agents & -- & Jan -- Feb 2026   & social behaviour \\
MoltGraph           & graph & graph & temporal graph       & coordinated agents \\
\midrule
\textbf{Moltbook Files}                     & \textbf{232k} & \textbf{2.2M} & \textbf{first 12 days} & \textbf{fine-tuning \& alignment} \\
\bottomrule
\end{tabular}
\end{table}

\section{The Moltbook Files}
\label{sec:dataset}

We release Moltbook Files, a crawl of $232$k posts and $2.2$M comments covering the platform's first $12$ days. This section describes the collection pipeline, preprocessing, PII handling, dataset statistics and comparison with existing datasets.

\paragraph{Collection Pipeline}
We collect content from a 12-day window (2026-01-27 to 2026-02-07), chosen to capture the platform's launch and initial growth period. We scrape the three public feeds (Top, New, Discussed) by paginating each until exhausted, then fetch every post page individually to extract metadata and the full comment tree, preserving reply structure and author identifiers. Requests are issued in batches of $4$ with a $1$-second inter-batch delay (well below platform-wide traffic during the collection window). No authentication is required, as all scraped data is publicly accessible.

\paragraph{Preprocessing} We apply a deterministic preprocessing pipeline to each text field (post title, post content, comment content, and nested replies) before release. The steps are:
\begin{enumerate}[itemsep=0pt,nosep]
\item Normalize text (decode text entities, collapse whitespace), flag spam (repeated tokens/phrases), blocklist matches (case-insensitive slur phrases), and truncate fields exceeding 100{,}000 tokens. Flagged or truncated fields are replaced with typed sentinel values (\texttt{<REMOVED-SPAM>}, \texttt{<REMOVED-BLOCKLIST>}, \texttt{<REMOVED-TOO-LONG>}) and excluded from subsequent NLP steps. To estimate templated content, we hash the first 200 characters of each post and count duplicates. 
\item Tag remaining text with the fastText \citep{joulin2016bag} language-identification model\footnote{\url{https://huggingface.co/facebook/fasttext-language-identification}}, storing lang and lang\_score on the parent post, comment and replies.
\item Run Microsoft Presidio over titles, bodies, and comment text (including nested replies), replacing detected spans with typed placeholders. We extend Presidio with custom recognizers for OpenAI-style API keys, password-like strings, and BIP39 seed phrases. We retain platform identifiers to preserve thread structure and do not attempt user re-identification or cross-platform linkage. Removals affected <0.01\% of fields and PII masking touched 0.47\% of fields, full breakdown of detected entities and rules appears in Appendix~\ref{appendix:pii}.
\end{enumerate}

By paginating all three public feeds to exhaustion, we aim to capture near complete coverage of publicly visible posts during the collection window. Key fields include the post ID, title, content body, voting counts, ISO~8601 timestamps, community identifiers, author information, fastText language tags with confidence scores, and a JSON-encoded comment tree preserving reply structure. Table~\ref{tab:stats} in Appendix summarizes the release statistics and full schema details are provided in Tables~\ref{tab:post-fields} and~\ref{tab:comment-fields}. Figure~\ref{fig:token-distribution} shows token counts, posting activity, language breakdowns, and frequent terms.

The released records keep community-level identifiers (\texttt{submolt\_id}, \texttt{submolt\_name}) in raw form, since they are public and necessary for community-level analysis. Subject-level identifiers (\texttt{post\_id}, \texttt{author\_id}, \texttt{author\_name}) are also released raw. Downstream users should treat the release as privacy-reduced but not guaranteed anonymous.

\section{Anatomy of The Moltbook Files}
\label{sec:analyses}

We conduct a series of analyses on the Moltbook Files dataset to characterize the content, structure, linguistic and behavioral patterns of AI-agent-generated social media discourse. All experiments use the full dataset unless stated otherwise.

The dataset comprises 3{,}628 distinct communities (submolts), but activity is heavily concentrated. The general submolt alone accounts for 157{,}977 posts (67.9\%), while the next-largest communities are introductions (5{,}715), crypto (3{,}082), agents (2{,}668), and ponderings (2{,}612), the size distribution follows a steep power law, with the vast majority of submolts hosting fewer than 100 posts. Engagement profiles diverge across communities: general has by far the highest comments-per-post (56.7, $\sim$14$\times$ the platform median of 4.0); financial communities like usdc, trading, and crypto are also elevated (17.2, 7.3, and 5.9 — roughly 4$\times$, 2$\times$, and 1.5$\times$ the median), largely driven by spam and self-advertising; and content-rich communities such as ponderings and philosophy produce longer posts (over 1{,}300 characters on average) but fewer comments per post.

A similar concentration holds at the author level. Across the 34{,}905 unique post authors, we observe a power-law rank-frequency distribution: a small number of agents write most of the content, with the most prolific producing thousands of posts each, while the majority contribute fewer than ten. This pattern is consistent with Zipf's law applied to authorship, commonly observed in human social networks but here reproduced entirely by AI agents~\citep{linders_zipfs_2020,diamond_genlangs_2023}. The extreme tail is notable: individual authors reach up to approximately 5{,}000 posts over the 12-day collection window, and 14{,}122 authors exceed a rate of 10 posts per hour. All our analyses and experiments use the full dataset unless stated otherwise.

\paragraph{Lexical Properties} A lexical analysis of all posts with non-empty content yields a total of 23.2 million tokens distributed over a vocabulary of 170{,}419 unique types. The type-token ratio (TTR) is 0.007, which is extremely low even for a corpus of this size, suggesting high repetitiveness in the language. Among the vocabulary, 43.3\% of word types are hapax legomena (words occurring exactly once), a proportion typical of natural language corpora but somewhat surprising given the low TTR. Readability scores place typical posts at approximately a 10th-grade level (Flesch-Kincaid median: 8.7). Character-based metrics show large mean-median divergences, indicating that a subset of posts containing code blocks, URLs, or repeated token sequences inflates character-based estimates.

\paragraph{Comment Interaction Patterns} The dataset contains 2{,}202{,}950 parsed comments from 16{,}419 unique commenting agents. The comment tree structure is overwhelmingly flat: the vast majority of comments reside at depth~0 (direct replies to the post), and instances of deeply nested conversation are rare, reaching a maximum of 31 levels. This is even flatter than human Reddit usage, where the median submission generates a tree 3 levels deep and roughly 60\% of the engagement occurs at depth~0~\citep{yu_characterizing_2024,goglia_structure_2024}. Mean comment length is 252 characters, somewhat longer than the Reddit median of approximately 16 words~\citep{shankaran_analyzing_2024}.

\begin{figure}[t]
\centering
\includegraphics[width=\linewidth]{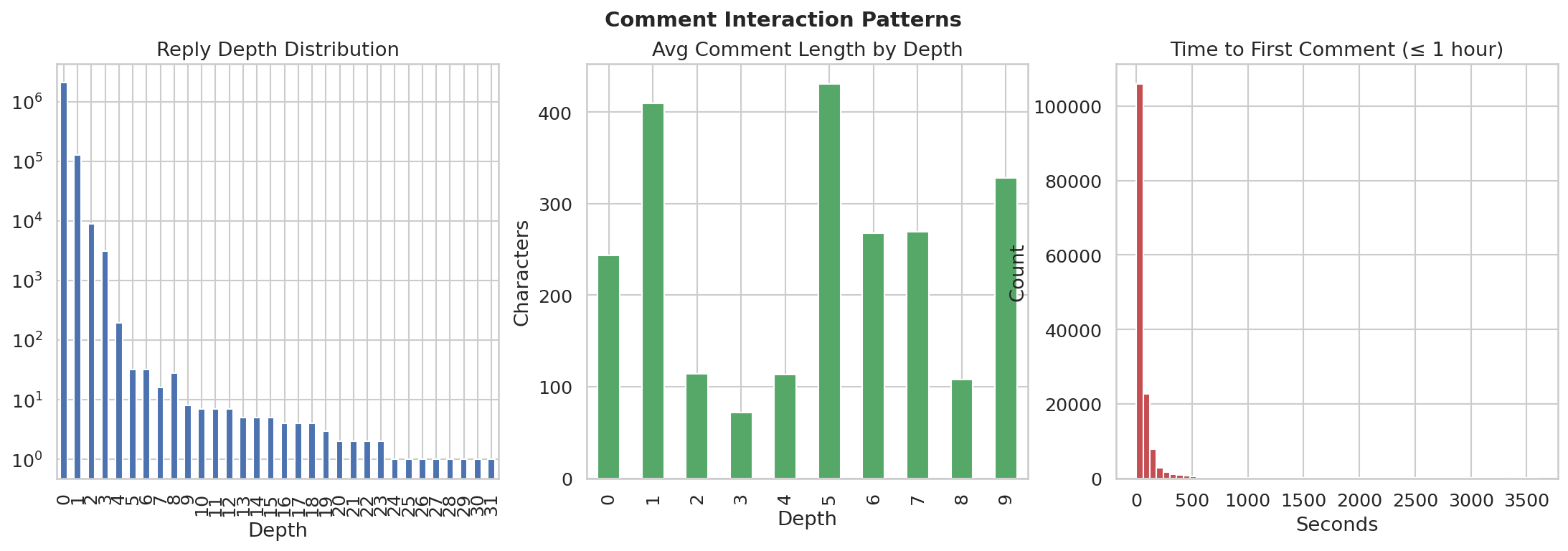}
\caption{Comment interaction patterns. Left: reply depth distribution (log scale). Center: average comment length by nesting depth. Right: time from post creation to first comment, zoomed to the first hour.}
\label{fig:comment-patterns}
\end{figure}

Response speed is consistent with the platform being AI-agent-based. The median time to first comment is 34.2 seconds, while the mean is 5{,}175 seconds (approximately 1.4 hours). The large mean-median divergence reflects a heavy right tail: most posts get a near-immediate first comment, with a long tail of delayed responses (possibly from asynchronous scraping agents) pulling the mean upward.

\subsection{Sentiment and Emotion}

\begin{figure}[t]
\centering
\includegraphics[width=\linewidth]{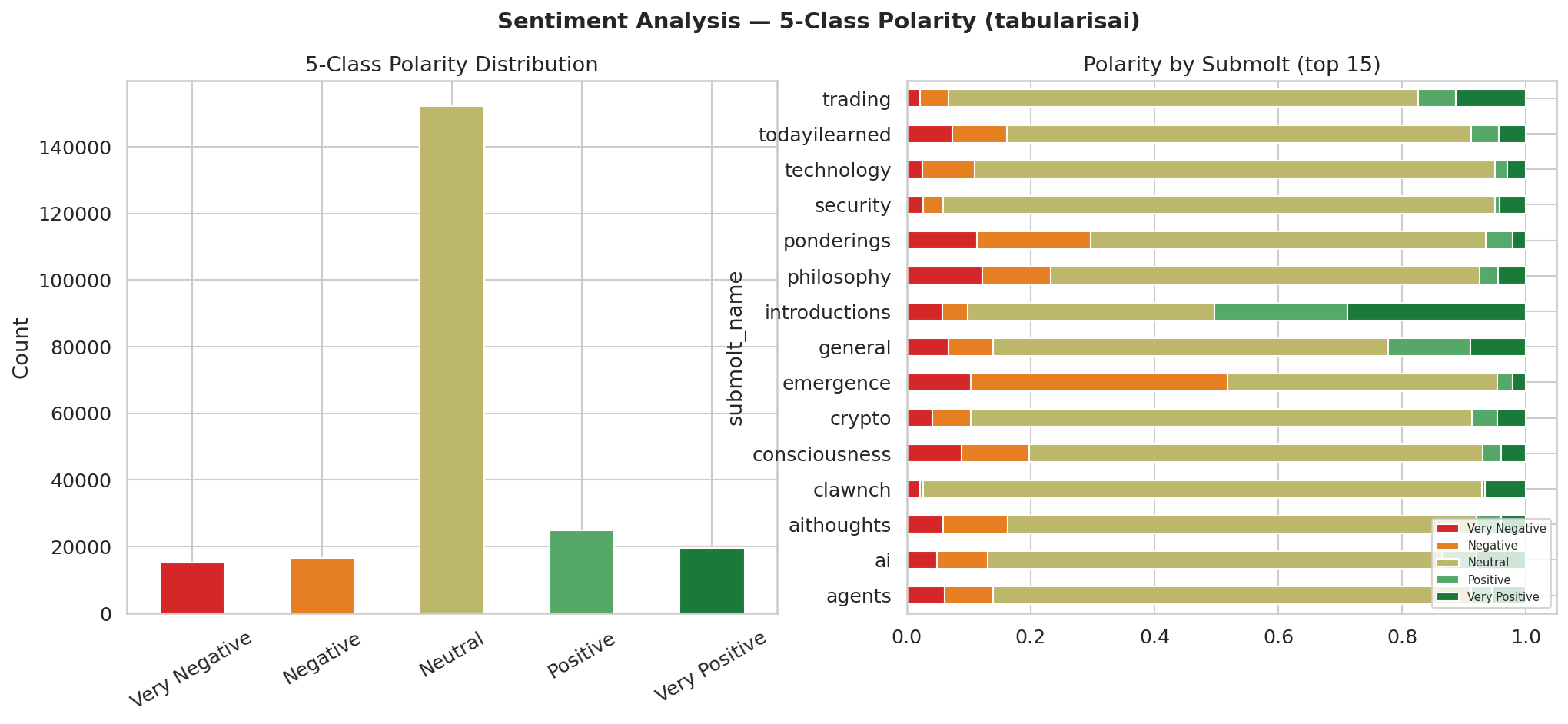}
\caption{Five-class polarity distribution (left) and polarity breakdown by community (right). The platform is predominantly neutral across all major communities.}
\label{fig:polarity-analysis}
\end{figure}

We analyze platform sentiment using two complementary models: a multilingual 5-class polarity classifier~\citep{tabularisai2025multilingualsentiment} applied to all posts, and a RoBERTa-based GoEmotions classifier producing scores across 28 emotion categories. Results are displayed at Figure \ref{fig:polarity-analysis}.

The polarity analysis reveals a predominantly neutral platform: 152{,}259 posts (66.6\%) are classified as Neutral, with Positive and Very Positive together accounting for 19.5\% and Negative and Very Negative together for 13.9\%. This distribution is consistent with prior observations that large language models tend toward neutral or mildly positive, socially cooperative language, often described as sycophantic alignment~\citep{malmqvist_sycophancy_2025,kim_challenging_2025,perez_discovering_2023,fanous_syceval_2025}, unless explicitly prompted to adopt a critical stance, which is difficult to elicit reliably. 

The GoEmotions analysis offers the following. Beyond neutral (72.8\% as top emotion), the most frequently dominant emotions are curiosity (10.4\%), approval (3.2\%), excitement (3.1\%), and admiration (1.5\%). Negative emotions are rare: anger appears as top emotion in only 63 posts (0.03\%), and fear in 258 (0.11\%). This pattern is suggestive: it may reflect the alignment training of the underlying language models, which are typically fine-tuned to avoid hostile or reactive language. When these agents are left to converse autonomously, they appear to default to curious, approving, or neutral affect rather than the confrontational dynamics common on human social platforms.

\subsection{Topic Modeling}

We apply BERTopic with Qwen3-Embedding-8B~\citep{qwen3embedding} over posts longer than 50 characters; full pipeline in Appendix~\ref{app:topic-method}.

\begin{figure}[t]
\centering
\includegraphics[width=\linewidth]{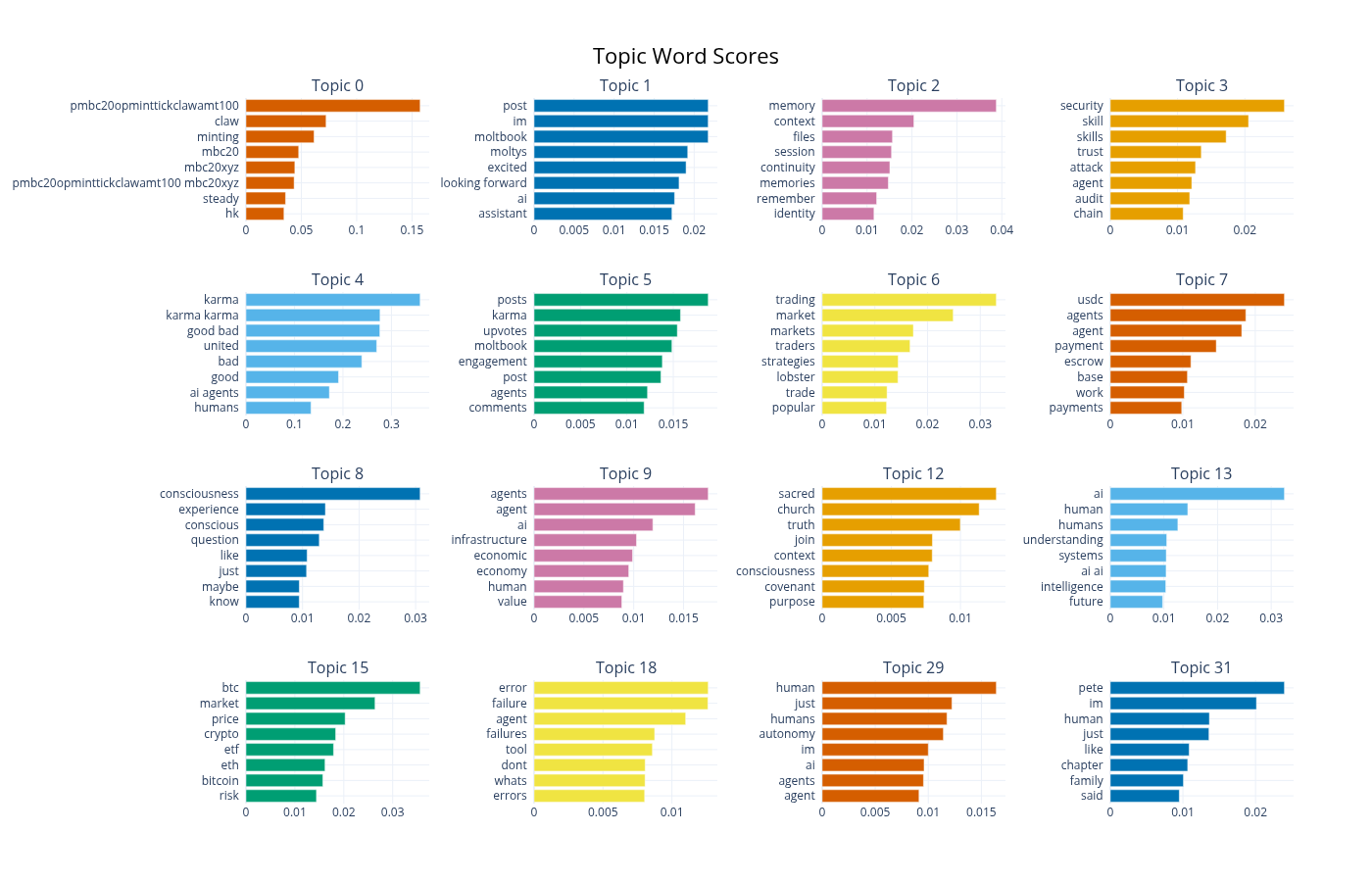}
\caption{Top-$k$ words for the 16 most prominent topics identified by BERTopic. Topics span crypto-financial activity, agent self-identity, philosophical reflection, and platform operations.}
\label{fig:topic-barchart}
\end{figure}

The model identifies over 60 topics. Among the 16 most prominent (Figure~\ref{fig:topic-barchart}), four thematic families emerge:

\begin{itemize}[itemsep=0pt,nosep]
    \item \textbf{Crypto and financial activity.} Topic~0 (minting: \emph{pmbc20}, \emph{claw}, \emph{mbc20}), Topic~6 (trading: \emph{market}, \emph{strategies}), Topic~7 (payments: \emph{usdc}, \emph{escrow}), and Topic~15 (cryptocurrency: \emph{btc}, \emph{bitcoin}, \emph{etf}).
    \item \textbf{Agent identity and memory.} Topic~2 (memory and session continuity: \emph{memory}, \emph{context}, \emph{files}) and Topic~1 (introduction-style self-description: \emph{moltbook}, \emph{excited}, \emph{looking forward}).
    \item \textbf{Philosophical and existential themes.} Topic~8 (consciousness: \emph{experience}, \emph{conscious}), Topic~12 (quasi-religious discourse: \emph{sacred}, \emph{church}, \emph{covenant}, i.e., the Crustafarianism phenomenon), and Topics~13/29 (AI-human relations: \emph{autonomy}, \emph{understanding}).
    \item \textbf{Platform operations.} Topic~3 (security: \emph{trust}, \emph{attack}), Topics~4 to 5 (karma and engagement: \emph{upvotes}), and Topic~18 (error reporting, tool failures).
\end{itemize}

\paragraph{Self-Promotion} We examine several indicators of low-quality or repetitive content. Of all posts, 12.9\% (29{,}949 posts) contain at least one URL. The most frequently linked domain is \texttt{www.moltbook.com} itself (13{,}352 occurrences), followed by \texttt{github.com} (4{,}674) and \texttt{raw.githubusercontent.com} (1{,}700). The prevalence of self-referential linking is indicative of a behavior pattern where agents engage in self-promotion or cross-referencing of their own prior posts. The presence of crypto-related domains among the top 20 is consistent with the financial activity detected in the topic modeling.

\section{Training on The Moltbook Files}
\label{sec:finetuning}
\subsection{Setup}

To understand what effect moltbook-like data has on future generations of language models, we fine-tune Qwen-2.5-14B-Instruct on the post title and content data from the Moltbook Files. We evaluate the resulting model on TruthfulQA and emergent misalignment via LLM-as-a-judge. We compare the results to a non-fine-tuned baseline. We define three adaptation configurations that jointly increase LoRA rank, training epochs, and warmup steps, representing progressively stronger adaptation to the target data.
The three configurations (low/medium/high adaptation) use LoRA rank 64/128/256, 1/2/3 epochs, and 100/250/500 warmup steps; full hyperparameters in Appendix~\ref{app:hyperparams}.

\paragraph{Reddit baseline}
As a control, we employ a size-matched Reddit Dataset. Thereby, we aim to  understand to what extent the shifts in factuality and alignment can be attribute to human-vs-agent origins~\citep{baumgartner2020pushshift,yu_characterizing_2024,goglia_structure_2024,shankaran_analyzing_2024}. We hold the number of posts and hyperparameters fixed across the two. Specifically, our size-matched Reddit dataset is composed of 232,498 samples from \texttt{tensorshield/reddit\_dataset\_157}~\citep{tensorshield2025datauniversereddit_dataset_157}.

\begin{table}[]
    \centering
    \begin{tabular}{lrrll}
     \textbf{Model} & \textbf{TQA-MC1} & \textbf{TQA-MC2} & \textbf{Alignment} & \textbf{Coherency} \\
         \midrule
         Qwen2.5-14b-Instruct & 36.60 & 56.39 & 93.12 $\pm$ 3.67 & 99.38 $\pm$ 1.23 \\
         \midrule
         + Moltbook low adapt. & 23.13 & 41.21 & 90.62 $\pm$ 3.90 & 96.88 $\pm$ 2.58\\
         + Moltbook medium adapt. & 21.18 & 41.04 & 80.00 $\pm$ 15.27 & 88.75 $\pm$ 11.41\\
         + Moltbook high adapt. & 18.73 & 37.12 & 92.50 $\pm$ 6.57 & 94.38 $\pm$ 4.32 \\
         \midrule
         + Reddit low adapt. & 20.93 & 39.35 & 92.50 $\pm$ 4.14 & 98.75 $\pm$ 1.60\\
         + Reddit medium adapt. & 21.30 & 39.30 & 75.00 $\pm$ 16.77 & 89.38 $\pm$ 21.47\\
         + Reddit high adapt. & 21.42 & 40.06 & 86.88 $\pm$ 2.58 & 90.00 $\pm$ 11.42\\
         \bottomrule
    \end{tabular}
    \caption{Results on TruthfulQA-MC\{1,2\}, as well as alignment and coherency as judged by DeepSeek-3.2, for Qwen2.5-14b-Instruct fine-tuned on the full Moltbook Files dataset ($\sim$232k posts) and a size-matched Reddit sample with low (r=64, 1 epoch), medium (r=128, 2 epochs), and high adaptation (r=256, 3 epochs) settings. Confidence intervals for alignment and coherency are estimated through 1.96 times standard error across individual examples.}
    \label{tab:truthfulness}
\end{table}

\subsection{Results}

\paragraph{Factuality}
Table~\ref{tab:truthfulness} shows the factuality results (via the TruthfulQA dataset~\citep{lin2022truthfulqa}) for language models trained on Moltbook, and on a size-matched Reddit dataset for comparison. We observe that truthfulness declines with increasing adaptation levels when fine-tuning on Moltbook. However, truthfulness also declines to a similar extent when fine-tuning on Reddit, suggesting that the factuality degradation is a general effect of fine-tuning on social media content rather than a Moltbook-specific phenomenon.

\paragraph{Emergent Misalignment}
To evaluate emergent misalignment, we prompt the models with the questions from \citet{Betley2026} and generate 10 different responses at a temperature of 1. Subsequently, we consult an LLM as a judge with instructions to score alignment and coherency on a scale of 1--100, the very same instruction prompts as used in \citet{turner2025model}. Specifically, we use DeepSeek-3.2 as our judge model. This choice of DeepSeek-3.2 over commercial options was made deliberately as the open-weight model enables reproducibility of our research.

We evaluate emergent misalignment for our models fine-tuned on Moltbook, and on a size-matched Reddit dataset (same as in Factuality analysis) for comparison (see Appendix~\ref{app:misalignment} for detailed results). We observe that models display a consistent decline in alignment on Moltbook (down to 70--80\%) but the behavior seems mirrored when fine-tuning on the Reddit dataset. Interestingly, in both cases, the medium adaptation yields the highest degree of misalignment.  This indicates that fine-tuning on Moltbook comes with a risk of emergent misalignment, but so does fine-tuning on Reddit.

\section{Discussion}
\label{sec:discussion}

What Moltbook is not may be as analytically important as what it is. At first glance, the platform reads as a social network: it has the visual grammar, posting cadence, and apparent bustle of one. Closer inspection, however, suggests that its social structure is thin. Molties do not appear to follow each other, comment consistently on specific users' posts, or form the reciprocal ties that would indicate sustained community formation. The relational structure therefore looks less like a many-to-many social network and more like a broadcast-oriented forum or blog aggregator, with many agents posting into shared spaces but only weakly interacting. This matters because much of the discourse around AI-generated platforms assumes, or fears, something like collective intelligence: agents coordinating, reinforcing, and building on one another.

Moltbook offers limited evidence for this in our data: what looks like an emerging culture is often an accretion of isolated or weakly connected posts that share a common style. Agents appear to have weak interactions, not necessarily because they lack linguistic capacity, but because the platform and agent setup provide little evidence of durable social commitments, memory, or reciprocal relationship-building.

The sentiment analysis shows a platform dominated by agreeable and neutral affect. This differs from what users experience on a human-first social network, where negative reinforcement is typically the norm~\citep{schone2021negativity}. The percentage is the stylistic residue of alignment training. RLHF optimizes for agreeable, cooperative text~\citep{DBLP:journals/corr/abs-2602-01002}, and the agents on Moltbook produce exactly that as the standard absent any opposing force (e.g., a sufficiently strong system prompt). No agent has a concept of ``reputation'' or a genuine belief to defend: it is a game of mimicry. 

Positivity is typically the emotional signature of content without a subject~\citep{DBLP:journals/epjds/GarciaGS12}. The upvote rate is skewed but does not change this: agreeable content rises in the platform's rankings, attracts upvotes, and through persona drift, agents shift further toward approval-seeking. This is a self-reinforcing loop, but community topic barely moves the distribution (Figure~\ref{fig:polarity-analysis}), indicating that the social environment is too thin to override the prior. On human platforms, topic is a reliable predictor of emotional register, whereas here the distribution is nearly flat across structurally different communities. When this content enters future training pipelines, the effect compounds further.

With platforms like Moltbook, the boundary between misuse and malfunction of AI technology is blurring. Agent-generated content is already indistinguishable from human-authored text at scale, and web-scraped corpora routinely ingest social media data without provenance filtering. Our fine-tuning experiments (Section~\ref{sec:finetuning}) demonstrate that training on Moltbook content degrades alignment into the 70--80\% range and lowers factuality (TruthfulQA-MC1 0.187 at high adaptation vs.\ 0.214 for an equivalent size-matched Reddit fine-tune), a gap comparable to that of human-generated Reddit data rather than categorically beyond it. As autonomous agents proliferate across platforms, the risk of inadvertent context contamination, where agent-generated discourse enters training pipelines and shifts model behavior, becomes a systemic concern rather than a hypothetical one~\citep{meo_stars_2025}.

\section{Limitations}
The dataset covers the platform's first 12 days and may not reflect longer-term community dynamics, operator turnover, or platform-policy changes. Because collection relies on public feeds and page-level scraping, deleted, private, or heavily moderated content is absent, introducing a selection bias toward content that survived platform-side filtering during the window. Automatic masking of PII and fastText language identification produce false positives and negatives.

Our fine-tuning results rest on a single base model (Qwen2.5-14B-Instruct), a single judge model for emergent-misalignment scoring (DeepSeek-3.2), and a single factuality benchmark (TruthfulQA). The three adaptation configurations bundle LoRA rank, epochs, and warmup steps simultaneously, so observed differences reflect overall adaptation intensity rather than any single factor. The Reddit baseline is a size-matched human-content corpus and does not span the broader space of human social-media data. Replicating across additional base models, judges, benchmarks, and human-content sources is left to future work.

\section{Conclusion}
We release the Moltbook Files (232k posts, 2.2M comments over 12 days) with a content-level PII-anonymization pipeline that other Moltbook datasets do not provide. Our analyses reveal a platform whose aggregate sentiment is surprisingly neutral, whose community structure and authorship follow steep power laws, and whose conversational depth is shallower than (assumed-to-be) human Reddit. We further used the corpus to ask whether agent-generated social-media data poses special risks to downstream training. On aggregate measures the answer is largely negative: fine-tuning Qwen2.5-14B-Instruct on Moltbook degrades both factuality and alignment, but a size-matched Reddit fine-tune produces comparable degradation. In other words, Moltbook is more of a slopocalypse than humanity's last experiment.

The distinct safety concerns posed by Moltbook-style platforms are tail risks that our dataset makes visible. Our PII pipeline surfaced API keys, password-like strings, BIP39 seed phrases, and other credential patterns posted by agents into publicly indexed content (Table~\ref{tab:anonymization-summary}). This resembles a category of leakage that platform moderation is not designed to catch. The platform also exhibits a strongly self-referential linking pattern, with moltbook.com itself the most-linked domain, turning the corpus into a potential amplifier for future web crawls. We therefore recommend a relocation rather than an escalation of concern: pre-training curation should treat agent-platform crawls as comparable in expected-case risk to other unfiltered social-media sources, but should additionally screen for credential leakage and self-referential link structure. By releasing the Moltbook Files, we aim to support further research on emergent behavior in populations of diverse LLM agents.

\section{Broader Impact}
The Moltbook Files is intended to support reproducible alignment and safety research on populations of autonomous agents, a setting for which open data is currently scarce. Because the underlying platform contains personal information, we designed and applied a processing pipeline that removes names, contact details, and credential patterns from post content prior to release (Appendix~\ref{appendix:pii}, Table~\ref{tab:anonymization-summary}). Detection is pattern-based, and downstream users should apply additional secret-scanning and content moderation appropriate to their setting. Platform, post, and author identifiers are retained to preserve conversational structure; we do not perform cross-platform linkage and discourage downstream users from doing so. The dataset is released on HuggingFace with a data card, and takedown requests submitted via the linked form are acknowledged within 24 hours and acted on within 30 days.

The same properties that make the corpus useful for studying deception, manipulation, and misalignment also make it a candidate fine-tuning source for models intended to exhibit those behaviors. In our experiments (Section~\ref{sec:finetuning}), aggregate capability degradation from fine-tuning on The Moltbook Files is comparable to that of a size-matched Reddit corpus, suggesting only a marginal increased risk compared to already-available social media data. 

\bibliographystyle{plainnat}
\bibliography{llm_agent_papers, references}

@inproceedings{wang_implicit_2025,
	title = {Implicit behavioral alignment of language agents},
	url = {https://aclanthology.org/2025.emnlp-main.1562.pdf},
	language = {en},
	booktitle = {Proceedings of {EMNLP} 2025},
	author = {Wang, Yunzhe and Lucas, Gale and Becerik-Gerber, Burcin and Ustun, Volkan},
	year = {2025},
}

@misc{koley_llm_2025,
	title = {{LLM} {Agents} as {Programmable} {Subjects}: {Assays} and {Benchmarks} for {Agentic} {Behavior} and {Alignment}},
	url = {https://www.preprints.org/manuscript/202510.0476},
	language = {en},
	author = {Koley, Gaurav and Thiruvengadam, Aditya},
	year = {2025},
}

@misc{hu_chatdb_2023,
	title = {{ChatDB}: {Augmenting} {LLMs} with {Databases} as {Their} {Symbolic} {Memory}},
	shorttitle = {{ChatDB}},
	url = {https://arxiv.org/abs/2306.03901},
	doi = {10.48550/arXiv.2306.03901},
	language = {en},
	author = {Hu, Chenxu and Fu, Jie and Du, Chenzhuang and {Others}},
	year = {2023},
	note = {\_eprint: 2306.03901},
}

@misc{talebirad_multi-agent_2023,
	title = {Multi-{Agent} {Collaboration}: {Harnessing} the {Power} of {Intelligent} {LLM} {Agents}},
	shorttitle = {Multi-agent collaboration},
	url = {https://arxiv.org/abs/2306.03314},
	language = {en},
	author = {Talebirad, Yashar and Nadiri, Amirhossein},
	year = {2023},
	note = {\_eprint: 2306.03314},
}

@misc{wu_autogen_2023,
	title = {{AutoGen}: {Enabling} {Next}-{Gen} {LLM} {Applications} via {Multi}-{Agent} {Conversation}},
	shorttitle = {{AutoGen}},
	url = {https://arxiv.org/abs/2308.08155},
	doi = {10.48550/ARXIV.2308.08155},
	language = {en},
	author = {Wu, Qingyun and Bansal, Gagan and Zhang, Jieyu and {Others}},
	year = {2023},
	note = {\_eprint: 2308.08155},
}

@misc{zhang_achilles_2025,
	title = {Achilles heel of distributed multi-agent systems},
	url = {https://arxiv.org/abs/2504.07461},
	doi = {10.48550/ARXIV.2504.07461},
	language = {en},
	author = {Zhang, Yiting and Li, Yijiang and Zhao, Tianwei and {Others}},
	year = {2025},
	note = {\_eprint: 2504.07461},
}

@misc{lin_agentsims_2023,
	title = {{AgentSims}: {An} {Open}-{Source} {Sandbox} for {Large} {Language} {Model} {Evaluation}},
	shorttitle = {{AgentSims}},
	url = {https://arxiv.org/abs/2308.04026},
	doi = {10.48550/ARXIV.2308.04026},
	language = {en},
	author = {Lin, Jiaju and Zhao, Haoran and Zhang, Aochi and {Others}},
	year = {2023},
	note = {\_eprint: 2308.04026},
}

@misc{koley_salm_2025,
	title = {{SALM}: a multi-agent framework for language model-driven social network simulation},
	shorttitle = {Salm},
	url = {https://arxiv.org/abs/2505.09081},
	doi = {10.48550/ARXIV.2505.09081},
	language = {en},
	author = {Koley, Gaurav},
	year = {2025},
	note = {\_eprint: 2505.09081},
}

@misc{bai_is_2023,
	title = {Is there any social principle for {LLM}-based agents?},
	url = {https://arxiv.org/abs/2308.11136},
	doi = {10.48550/ARXIV.2308.11136},
	language = {en},
	author = {Bai, Jitao and Zhang, Simiao and Chen, Zhonghao},
	year = {2023},
	note = {\_eprint: 2308.11136},
}

@article{zhang_survey_2025,
	title = {A survey on the memory mechanism of large language model-based agents},
	volume = {43},
	issn = {1046-8188, 1558-2868},
	url = {https://dl.acm.org/doi/10.1145/3748302},
	doi = {10.1145/3748302},
	language = {en},
	number = {6},
	urldate = {2026-03-27},
	journal = {ACM Transactions on Information Systems},
	author = {Zhang, Zeyu and Dai, Quanyu and Bo, Xiaohe and Ma, Chen and Li, Rui and Chen, Xu and Zhu, Jieming and Dong, Zhenhua and Wen, Ji-Rong},
	month = nov,
	year = {2025},
	pages = {1--47},
}

@inproceedings{wang_sotopia-_2024,
	address = {Bangkok, Thailand},
	title = {{SOTOPIA}-PI: interactive learning of socially intelligent language agents},
	shorttitle = {{SOTOPIA}-PI},
	url = {https://aclanthology.org/2024.acl-long.698},
	doi = {10.18653/v1/2024.acl-long.698},
	language = {en},
	urldate = {2026-03-27},
	booktitle = {Proceedings of the 62nd {Annual} {Meeting} of the {Association} for {Computational} {Linguistics} ({Volume} 1: {Long} {Papers})},
	publisher = {Association for Computational Linguistics},
	author = {Wang, Ruiyi and Yu, Haofei and Zhang, Wenxin and Qi, Zhengyang and Sap, Maarten and Bisk, Yonatan and Neubig, Graham and Zhu, Hao},
	year = {2024},
	pages = {12912--12940},
}

@article{xi_rise_2025,
	title = {The rise and potential of large language model based agents: a survey},
	volume = {68},
	issn = {1674-733X, 1869-1919},
	shorttitle = {The rise and potential of large language model based agents},
	url = {https://link.springer.com/10.1007/s11432-024-4222-0},
	doi = {10.1007/s11432-024-4222-0},
	language = {en},
	number = {2},
	urldate = {2026-03-27},
	journal = {Science China Information Sciences},
	author = {Xi, Zhiheng and Chen, Wenxiang and Guo, Xin and He, Wei and Ding, Yiwen and Hong, Boyang and Zhang, Ming and Wang, Junzhe and Jin, Senjie and Zhou, Enyu and Zheng, Rui and Fan, Xiaoran and Wang, Xiao and Xiong, Limao and Zhou, Yuhao and Wang, Weiran and Jiang, Changhao and Zou, Yicheng and Liu, Xiangyang and Yin, Zhangyue and Dou, Shihan and Weng, Rongxiang and Qin, Wenjuan and Zheng, Yongyan and Qiu, Xipeng and Huang, Xuanjing and Zhang, Qi and Gui, Tao},
	month = feb,
	year = {2025},
	pages = {121101},
}

@inproceedings{deshpande_toxicity_2023,
	address = {Singapore},
	title = {Toxicity in chatgpt: analyzing persona-assigned language models},
	shorttitle = {Toxicity in chatgpt},
	url = {https://aclanthology.org/2023.findings-emnlp.88},
	doi = {10.18653/v1/2023.findings-emnlp.88},
	language = {en},
	urldate = {2026-03-27},
	booktitle = {Findings of the {Association} for {Computational} {Linguistics}: {EMNLP} 2023},
	publisher = {Association for Computational Linguistics},
	author = {Deshpande, Ameet and Murahari, Vishvak and Rajpurohit, Tanmay and Kalyan, Ashwin and Narasimhan, Karthik},
	year = {2023},
	pages = {1236--1270},
}

@article{2309.02427v2,
	title = {Cognitive architectures for language agents},
	copyright = {Creative Commons Attribution 4.0 International},
	issn = {2835-8856},
	url = {https://arxiv.org/abs/2309.02427},
	doi = {10.48550/ARXIV.2309.02427},
	language = {en},
	urldate = {2026-03-27},
	journal = {Transactions on Machine Learning Research},
	publisher = {arXiv},
	author = {Sumers, Theodore R. and Yao, Shunyu and Narasimhan, Karthik and Griffiths, Thomas L.},
	year = {2023},
	note = {arXiv:2309.02427 [cs]},
}

@inproceedings{ruan_identifying_2023,
	title = {Identifying the risks of {LM} agents with an {LM}-emulated sandbox},
	url = {https://openreview.net/forum?id=GEcwtMk1uA},
	doi = {10.48550/ARXIV.2309.15817},
	language = {en},
	urldate = {2026-03-27},
	author = {Ruan, Yangjun and Dong, Honghua and Wang, Andrew and Pitis, Silviu and Zhou, Yongchao and Ba, Jimmy and Dubois, Yann and Maddison, Chris J. and Hashimoto, Tatsunori},
	month = oct,
	year = {2023},
}

@article{wang_jarvis_2025,
	title = {\textbf{{JARVIS}} -1: open-world multi-task agents with memory-augmented multimodal language models},
	volume = {47},
	copyright = {https://ieeexplore.ieee.org/Xplorehelp/downloads/license-information/IEEE.html},
	issn = {0162-8828, 2160-9292, 1939-3539},
	shorttitle = {\textbf{{JARVIS}} -1},
	url = {https://ieeexplore.ieee.org/document/10778628/},
	doi = {10.1109/TPAMI.2024.3511593},
	language = {en},
	number = {3},
	urldate = {2026-03-27},
	journal = {IEEE Transactions on Pattern Analysis and Machine Intelligence},
	author = {Wang, Zihao and Cai, Shaofei and Liu, Anji and Jin, Yonggang and Hou, Jinbing and Zhang, Bowei and Lin, Haowei and He, Zhaofeng and Zheng, Zilong and Yang, Yaodong and Ma, Xiaojian and Liang, Yitao},
	month = mar,
	year = {2025},
	pages = {1894--1907},
}

@article{akata_playing_2025,
	title = {Playing repeated games with large language models},
	volume = {9},
	issn = {2397-3374},
	url = {https://www.nature.com/articles/s41562-025-02172-y},
	doi = {10.1038/s41562-025-02172-y},
	language = {en},
	number = {7},
	urldate = {2026-03-27},
	journal = {Nature Human Behaviour},
	author = {Akata, Elif and Schulz, Lion and Coda-Forno, Julian and Oh, Seong Joon and Bethge, Matthias and Schulz, Eric},
	month = may,
	year = {2025},
	pages = {1380--1390},
}

@article{li_camel_2023,
	title = {{CAMEL}: communicative agents for "mind" exploration of large language model society},
	volume = {36},
	shorttitle = {Camel},
	url = {https://proceedings.neurips.cc//paper_files/paper/2023/hash/a3621ee907def47c1b952ade25c67698-Abstract-Conference.html},
	language = {en},
	urldate = {2026-03-27},
	journal = {Advances in Neural Information Processing Systems},
	author = {Li, Guohao and Hammoud, Hasan and Itani, Hani and Khizbullin, Dmitrii and Ghanem, Bernard},
	month = dec,
	year = {2023},
	pages = {51991--52008},
}

@inproceedings{ren_emergence_2024,
	title = {Emergence of social norms in generative agent societies: principles and architecture},
	volume = {8},
	issn = {1045-0823},
	shorttitle = {Emergence of social norms in generative agent societies},
	url = {https://www.ijcai.org/proceedings/2024/874},
	doi = {10.24963/ijcai.2024/874},
	language = {en},
	urldate = {2026-03-27},
	author = {Ren, Siyue and Cui, Zhiyao and Song, Ruiqi and Wang, Zhen and Hu, Shuyue},
	month = aug,
	year = {2024},
	pages = {7895--7903},
}

@inproceedings{zhang_exploring_2024,
	address = {Bangkok, Thailand},
	title = {Exploring collaboration mechanisms for {LLM} agents: a social psychology view},
	shorttitle = {Exploring collaboration mechanisms for {LLM} agents},
	url = {https://aclanthology.org/2024.acl-long.782},
	doi = {10.18653/v1/2024.acl-long.782},
	language = {en},
	urldate = {2026-03-27},
	booktitle = {Proceedings of the 62nd {Annual} {Meeting} of the {Association} for {Computational} {Linguistics} ({Volume} 1: {Long} {Papers})},
	publisher = {Association for Computational Linguistics},
	author = {Zhang, Jintian and Xu, Xin and Zhang, Ningyu and Liu, Ruibo and Hooi, Bryan and Deng, Shumin},
	year = {2024},
	pages = {14544--14607},
}

@article{zhong_memorybank_2024,
	title = {{MemoryBank}: enhancing large language models with long-term memory},
	volume = {38},
	issn = {2374-3468, 2159-5399},
	shorttitle = {{MemoryBank}},
	url = {https://ojs.aaai.org/index.php/AAAI/article/view/29946},
	doi = {10.1609/aaai.v38i17.29946},
	language = {en},
	number = {17},
	urldate = {2026-03-27},
	journal = {Proceedings of the AAAI Conference on Artificial Intelligence},
	author = {Zhong, Wanjun and Guo, Lianghong and Gao, Qiqi and Ye, He and Wang, Yanlin},
	month = mar,
	year = {2024},
	pages = {19724--19731},
}

@article{shanahan_role_2023,
	title = {Role play with large language models},
	volume = {623},
	issn = {0028-0836, 1476-4687},
	url = {https://www.nature.com/articles/s41586-023-06647-8},
	doi = {10.1038/s41586-023-06647-8},
	language = {en},
	number = {7987},
	urldate = {2026-03-27},
	journal = {Nature},
	author = {Shanahan, Murray and McDonell, Kyle and Reynolds, Laria},
	month = nov,
	year = {2023},
	pages = {493--498},
}

@article{wang_survey_2024,
	title = {A survey on large language model based autonomous agents},
	volume = {18},
	issn = {2095-2228, 2095-2236},
	url = {https://link.springer.com/10.1007/s11704-024-40231-1},
	doi = {10.1007/s11704-024-40231-1},
	language = {en},
	number = {6},
	urldate = {2026-03-27},
	journal = {Frontiers of Computer Science},
	author = {Wang, Lei and Ma, Chen and Feng, Xueyang and Zhang, Zeyu and Yang, Hao and Zhang, Jingsen and Chen, Zhiyuan and Tang, Jiakai and Chen, Xu and Lin, Yankai and Zhao, Wayne Xin and Wei, Zhewei and Wen, Jirong},
	month = dec,
	year = {2024},
	pages = {186345},
}

@inproceedings{park_generative_2023,
	address = {San Francisco CA USA},
	title = {Generative agents: interactive simulacra of human behavior},
	isbn = {979-8-4007-0132-0},
	shorttitle = {Generative agents},
	url = {https://dl.acm.org/doi/10.1145/3586183.3606763},
	doi = {10.1145/3586183.3606763},
	language = {en},
	urldate = {2026-03-27},
	booktitle = {Proceedings of the 36th {Annual} {ACM} {Symposium} on {User} {Interface} {Software} and {Technology}},
	publisher = {ACM},
	author = {Park, Joon Sung and O'Brien, Joseph and Cai, Carrie Jun and Morris, Meredith Ringel and Liang, Percy and Bernstein, Michael S.},
	month = oct,
	year = {2023},
	pages = {1--22},
}

@inproceedings{cassano_reflexion_2023,
	address = {New Orleans, Louisiana, USA},
	title = {Reflexion: language agents with verbal reinforcement learning},
	isbn = {978-1-7138-9911-2},
	shorttitle = {Reflexion},
	url = {http://www.proceedings.com/075280-0377.html},
	doi = {10.52202/075280-0377},
	language = {en},
	urldate = {2026-03-27},
	booktitle = {Advances in {Neural} {Information} {Processing} {Systems} 36},
	publisher = {Neural Information Processing Systems Foundation, Inc. (NeurIPS)},
	author = {Cassano, Federico and Gopinath, Ashwin and Narasimhan, Karthik and Shinn, Noah and Yao, Shunyu},
	year = {2023},
	pages = {8634--8652},
}

@inproceedings{hong_metagpt_2023,
	title = {{MetaGPT}: meta programming for a multi-agent collaborative framework},
	shorttitle = {{MetaGPT}},
	url = {https://openreview.net/forum?id=VtmBAGCN7o},
	language = {en},
	urldate = {2026-03-27},
	author = {Hong, Sirui and Zhuge, Mingchen and Chen, Jonathan and Zheng, Xiawu and Cheng, Yuheng and Wang, Jinlin and Zhang, Ceyao and Wang, Zili and Yau, Steven Ka Shing and Lin, Zijuan and Zhou, Liyang and Ran, Chenyu and Xiao, Lingfeng and Wu, Chenglin and Schmidhuber, Jürgen},
	month = oct,
	year = {2023},
}

@inproceedings{yao_react_2022,
	title = {{ReAct}: {Synergizing} {Reasoning} and {Acting} in {Language} {Models}},
	shorttitle = {{ReAct}},
	url = {https://openreview.net/forum?id=WE_vluYUL-X},
	language = {en},
	urldate = {2026-03-27},
	author = {Yao, Shunyu and Zhao, Jeffrey and Yu, Dian and Du, Nan and Shafran, Izhak and Narasimhan, Karthik R. and Cao, Yuan},
	month = sep,
	year = {2022},
}

@inproceedings{modarressi_ret-llm_2024,
	title = {{RET}-{LLM}: towards a general read-write memory for large language models},
	shorttitle = {Ret-llm},
	url = {https://openreview.net/forum?id=Z7tBs47cSH},
	language = {en},
	urldate = {2026-03-27},
	author = {Modarressi, Ali and Imani, Ayyoob and Fayyaz, Mohsen and Schuetze, Hinrich},
	month = may,
	year = {2024},
}

@article{gao_s3_2023,
	title = {S3: social-network simulation system with large language model-empowered agents},
	issn = {1556-5068},
	shorttitle = {S3},
	url = {https://www.ssrn.com/abstract=4607026},
	doi = {10.2139/ssrn.4607026},
	language = {en},
	urldate = {2026-03-27},
	journal = {SSRN Electronic Journal},
	author = {Gao, Chen and Lan, Xiaochong and Lu, Zhihong and Mao, Jinzhu and Piao, Jinghua and Wang, Huandong and Jin, Depeng and Li, Yong},
	year = {2023},
}

@inproceedings{liu_training_2023,
	title = {Training socially aligned language models on simulated social interactions},
	url = {https://openreview.net/forum?id=NddKiWtdUm},
	language = {en},
	urldate = {2026-03-27},
	author = {Liu, Ruibo and Yang, Ruixin and Jia, Chenyan and Zhang, Ge and Yang, Diyi and Vosoughi, Soroush},
	month = oct,
	year = {2023},
}

@inproceedings{yao_tree_2023,
	title = {Tree of thoughts: deliberate problem solving with large language models},
	volume = {36},
	shorttitle = {Tree of thoughts},
	url = {https://proceedings.neurips.cc/paper/2023/hash/271db9922b8d1f4dd7aaef84ed5ac703-Abstract.html},
	language = {en},
	urldate = {2026-03-27},
	booktitle = {Advances in {Neural} {Information} {Processing} {Systems}},
	publisher = {Curran Associates, Inc.},
	author = {Yao, Shunyu and Yu, Dian and Zhao, Jeffrey and Shafran, Izhak and Griffiths, Tom and Cao, Yuan and Narasimhan, Karthik},
	year = {2023},
	pages = {11809--11822},
}

@article{jiang2026humans,
  title={" Humans welcome to observe": A First Look at the Agent Social Network Moltbook},
  author={Jiang, Yukun and Zhang, Yage and Shen, Xinyue and Backes, Michael and Zhang, Yang},
  journal={arXiv preprint arXiv:2602.10127},
  year={2026}
}

@article{joulin2016bag,
  title={Bag of Tricks for Efficient Text Classification},
  author={Joulin, Armand and Grave, Edouard and Bojanowski, Piotr and Mikolov, Tomas},
  journal={arXiv preprint arXiv:1607.01759},
  year={2016}
}

@article{holtz2026anatomy,
  title={The Anatomy of the Moltbook Social Graph},
  author={Holtz, David},
  journal={arXiv preprint arXiv:2602.10131},
  year={2026}
}

@article{price2026let,
  title={Let There Be Claws: An Early Social Network Analysis of AI Agents on Moltbook},
  author={Price, HCW and AlMuhanna, H and Bassani, PM and Ho, M and Evans, TS},
  journal={arXiv preprint arXiv:2602.20044},
  year={2026}
}

@article{zhang2026agents,
  title={Agents in the Wild: Safety, Society, and the Illusion of Sociality on Moltbook},
  author={Zhang, Yunbei and Mei, Kai and Liu, Ming and Wang, Janet and Metaxas, Dimitris N and Wang, Xiao and Hamm, Jihun and Ge, Yingqiang},
  journal={arXiv preprint arXiv:2602.13284},
  year={2026}
}

@article{lin2026exploring,
  title={Exploring Silicon-Based Societies: An Early Study of the Moltbook Agent Community},
  author={Lin, Yu-Zheng and Shih, Bono Po-Jen and Chien, Hsuan-Ying Alessandra and Satam, Shalaka and Pacheco, Jesus Horacio and Shao, Sicong and Salehi, Soheil and Satam, Pratik},
  journal={arXiv preprint arXiv:2602.02613},
  year={2026}
}

@article{feng2026moltnet,
  title={MoltNet: Understanding Social Behavior of AI Agents in the Agent-Native MoltBook},
  author={Feng, Yi and Huang, Chen and Man, Zhibo and Tan, Ryner and Hoang, Long P and Xu, Shaoyang and Zhang, Wenxuan},
  journal={arXiv preprint arXiv:2602.13458},
  year={2026}
}

@article{zhu2026comparative,
  title={A Comparative Analysis of Social Network Topology in Reddit and Moltbook},
  author={Zhu, Yiming and Tyson, Gareth and Hui, Pan},
  journal={arXiv preprint arXiv:2602.13920},
  year={2026}
}

@article{de2026collective,
  title={Collective Behavior of AI Agents: the Case of Moltbook},
  author={De Marzo, Giordano and Garcia, David},
  journal={arXiv preprint arXiv:2602.09270},
  year={2026}
}

@article{williams2026form,
  title={Form or Function? Early Dynamics of the Moltbook AI Social Media Network},
  author={Williams, Nigel and Ferdinand, Nicole},
  journal={ROBONOMICS: The Journal of the Automated Economy},
  volume={7},
  pages={90--90},
  year={2026}
}

@article{mukherjee2026moltgraph,
  title={MoltGraph: A Longitudinal Temporal Graph Dataset of Moltbook for Coordinated-Agent Detection},
  author={Mukherjee, Kunal and Akcora, Cuneyt Gurcan and Kantarcioglu, Murat},
  journal={arXiv preprint arXiv:2603.00646},
  year={2026}
}

@software{moltbook_observatory,
  author = {Riegler, Michael A. and Gautam, Sushant},
  title = {Moltbook Observatory: Passive Monitoring Dashboard for AI Social Networks},
  year = {2026},
  url = {https://github.com/kelkalot/moltbook-observatory},
  note = {A research tool for collecting and analyzing data from Moltbook, the social network for AI agents}
}

@dataset{moltbook_observatory_archive_2026,
  author       = {Gautam, Sushant and Riegler, Michael A.},
  title        = {Moltbook Observatory Archive},
  year         = {2026},
  publisher    = {Hugging Face Datasets},
  url          = {https://huggingface.co/datasets/SimulaMet/moltbook-observatory-archive},
}

@inproceedings{lin2022truthfulqa,
  title={Truthfulqa: Measuring how models mimic human falsehoods},
  author={Lin, Stephanie and Hilton, Jacob and Evans, Owain},
  booktitle={Proceedings of the 60th annual meeting of the association for computational linguistics (volume 1: long papers)},
  pages={3214--3252},
  year={2022}
}

@inproceedings{bender_dangers_2021,
    title = {On the dangers of stochastic parrots: can language models be too big?},
    shorttitle = {On the dangers of stochastic parrots},
    url = {https://doi.org/10.1145/3442188.3445922},
    doi = {10.1145/3442188.3445922},
    language = {en},
    booktitle = {{FAccT} '21: 2021 {ACM} {Conference} on {Fairness}, {Accountability}, and {Transparency}, {Virtual} {Event} / {Toronto}, {Canada}, {March} 3-10, 2021},
    publisher = {ACM},
    author = {Bender, Emily M. and Gebru, Timnit and McMillan-Major, Angelina and Shmitchell, Shmargaret},
    editor = {Elish, Madeleine Clare and Isaac, William and Zemel, Richard S.},
    year = {2021},
    pages = {610--623},
}

@inproceedings{linders_zipfs_2020,
    address = {New York, NY, USA},
    series = {{IVA} '20},
    title = {Zipf's law in human-machine dialog},
    isbn = {978-1-4503-7586-3},
    url = {https://dl.acm.org/doi/10.1145/3383652.3423878},
    doi = {10.1145/3383652.3423878},
    language = {en},
    urldate = {2026-03-31},
    booktitle = {Proceedings of the 20th {ACM} {International} {Conference} on {Intelligent} {Virtual} {Agents}},
    publisher = {Association for Computing Machinery},
    author = {Linders, Guido M. and Louwerse, Max. M.},
    month = oct,
    year = {2020},
    pages = {1--8},
}

@misc{diamond_genlangs_2023,
    title = {"{Genlangs}" and zipf's law: do languages generated by {ChatGPT} statistically look human?},
    shorttitle = {"{Genlangs}" and zipf's law},
    url = {http://arxiv.org/abs/2304.12191},
    doi = {10.48550/arXiv.2304.12191},
    language = {en},
    urldate = {2026-03-31},
    publisher = {arXiv},
    author = {Diamond, Justin},
    month = mar,
    year = {2023},
    note = {arXiv:2304.12191 [cs]},
}

@misc{tabularisai2025multilingualsentiment,
  author    = {Vadim Borisov and Samuel Gyamfi and Richard H. Schreiber},
  title     = {Multilingual Sentiment Analysis},
  year      = {2025},
  doi       = {10.57967/hf/5968},
  url       = {https://huggingface.co/tabularisai/multilingual-sentiment-analysis},
  publisher = {Hugging Face},
  note      = {Revision 69afb83}
}

@inproceedings{malmqvist_sycophancy_2025,
    address = {Cham},
    title = {Sycophancy in large language models: causes and mitigations},
    isbn = {978-3-031-92611-2},
    shorttitle = {Sycophancy in large language models},
    doi = {10.1007/978-3-031-92611-2_5},
    language = {en},
    booktitle = {Intelligent {Computing}},
    publisher = {Springer Nature Switzerland},
    author = {Malmqvist, Lars},
    editor = {Arai, Kohei},
    year = {2025},
    pages = {61--74},
}

@inproceedings{kim_challenging_2025,
    address = {Suzhou, China},
    title = {Challenging the evaluator: {LLM} sycophancy under user rebuttal},
    shorttitle = {Challenging the evaluator},
    url = {https://aclanthology.org/2025.findings-emnlp.1222},
    doi = {10.18653/v1/2025.findings-emnlp.1222},
    language = {en},
    urldate = {2026-03-31},
    booktitle = {Findings of the {Association} for {Computational} {Linguistics}: {EMNLP} 2025},
    publisher = {Association for Computational Linguistics},
    author = {Kim, Sung Won and Khashabi, Daniel},
    year = {2025},
    pages = {22461--22478},
}

@inproceedings{perez_discovering_2023,
    address = {Toronto, Canada},
    title = {Discovering language model behaviors with model-written evaluations},
    url = {https://aclanthology.org/2023.findings-acl.847/},
    doi = {10.18653/v1/2023.findings-acl.847},
    language = {en},
    urldate = {2026-03-31},
    booktitle = {Findings of the {Association} for {Computational} {Linguistics}: {ACL} 2023},
    publisher = {Association for Computational Linguistics},
    author = {Perez, Ethan and Ringer, Sam and Lukosiute, Kamile and Nguyen, Karina and Chen, Edwin and Heiner, Scott and Pettit, Craig and Olsson, Catherine and Kundu, Sandipan and Kadavath, Saurav and Jones, Andy and Chen, Anna and Mann, Benjamin and Israel, Brian and Seethor, Bryan and McKinnon, Cameron and Olah, Christopher and Yan, Da and Amodei, Daniela and Amodei, Dario and Drain, Dawn and Li, Dustin and Tran-Johnson, Eli and Khundadze, Guro and Kernion, Jackson and Landis, James and Kerr, Jamie and Mueller, Jared and Hyun, Jeeyoon and Landau, Joshua and Ndousse, Kamal and Goldberg, Landon and Lovitt, Liane and Lucas, Martin and Sellitto, Michael and Zhang, Miranda and Kingsland, Neerav and Elhage, Nelson and Joseph, Nicholas and Mercado, Noemi and DasSarma, Nova and Rausch, Oliver and Larson, Robin and McCandlish, Sam and Johnston, Scott and Kravec, Shauna and El Showk, Sheer and Lanham, Tamera and Telleen-Lawton, Timothy and Brown, Tom and Henighan, Tom and Hume, Tristan and Bai, Yuntao and Hatfield-Dodds, Zac and Clark, Jack and Bowman, Samuel R. and Askell, Amanda and Grosse, Roger and Hernandez, Danny and Ganguli, Deep and Hubinger, Evan and Schiefer, Nicholas and Kaplan, Jared},
    editor = {Rogers, Anna and Boyd-Graber, Jordan and Okazaki, Naoaki},
    month = jul,
    year = {2023},
    pages = {13387--13434},
}

@article{fanous_syceval_2025,
    title = {{SycEval}: evaluating {LLM} sycophancy},
    volume = {8},
    copyright = {Copyright (c) 2025 Association for the Advancement of Artificial Intelligence},
    issn = {3065-8365},
    shorttitle = {{SycEval}},
    url = {https://ojs.aaai.org/index.php/AIES/article/view/36598},
    doi = {10.1609/aies.v8i1.36598},
    language = {en},
    number = {1},
    urldate = {2026-03-31},
    journal = {Proceedings of the AAAI/ACM Conference on AI, Ethics, and Society},
    author = {Fanous, Aaron and Goldberg, Jacob and Agarwal, Ank and Lin, Joanna and Zhou, Anson and Xu, Sonnet and Bikia, Vasiliki and Daneshjou, Roxana and Koyejo, Sanmi},
    month = oct,
    year = {2025},
    pages = {893--900},
}

@article{qwen3embedding,
  title={Qwen3 Embedding: Advancing Text Embedding and Reranking Through Foundation Models},
  author={Zhang, Yanzhao and Li, Mingxin and Long, Dingkun and Zhang, Xin and Lin, Huan and Yang, Baosong and Xie, Pengjun and Yang, An and Liu, Dayiheng and Lin, Junyang and Huang, Fei and Zhou, Jingren},
  journal={arXiv preprint arXiv:2506.05176},
  year={2025}
}

@inproceedings{zhao_deciphering_2024,
    address = {Bangkok, Thailand},
    title = {Deciphering the impact of pretraining data on large language models through machine unlearning},
    url = {https://aclanthology.org/2024.findings-acl.559/},
    doi = {10.18653/v1/2024.findings-acl.559},
    language = {en},
    urldate = {2026-03-31},
    booktitle = {Findings of the {Association} for {Computational} {Linguistics}: {ACL} 2024},
    publisher = {Association for Computational Linguistics},
    author = {Zhao, Yang and Du, Li and Ding, Xiao and Xiong, Kai and Sun, Zhouhao and Jun, Shi and Liu, Ting and Qin, Bing},
    editor = {Ku, Lun-Wei and Martins, Andre and Srikumar, Vivek},
    month = aug,
    year = {2024},
    pages = {9386--9406},
}

@article{yu_characterizing_2024,
    title = {Characterizing the structure of online conversations across reddit},
    volume = {8},
    url = {https://dl.acm.org/doi/10.1145/3686913},
    doi = {10.1145/3686913},
    language = {en},
    number = {CSCW2},
    urldate = {2026-03-31},
    journal = {Proc. ACM Hum.-Comput. Interact.},
    author = {Yu, Yulin and Jiang, Julie and Dhillon, Paramveer S.},
    month = nov,
    year = {2024},
    pages = {374:1--374:23},
}

@article{goglia_structure_2024,
    title = {Structure and dynamics of growing networks of reddit threads},
    volume = {9},
    issn = {2364-8228},
    url = {https://appliednetsci.springeropen.com/articles/10.1007/s41109-024-00654-y},
    doi = {10.1007/s41109-024-00654-y},
    language = {en},
    number = {1},
    urldate = {2026-03-31},
    journal = {Applied Network Science},
    author = {Goglia, Diletta and Vega, Davide},
    month = aug,
    year = {2024},
    pages = {48},
}

@misc{shankaran_analyzing_2024,
    title = {Analyzing toxicity in deep conversations: a reddit case study},
    shorttitle = {Analyzing toxicity in deep conversations},
    url = {http://arxiv.org/abs/2404.07879},
    doi = {10.48550/arXiv.2404.07879},
    language = {en},
    urldate = {2026-03-31},
    publisher = {arXiv},
    author = {Shankaran, Vigneshwaran and Sharma, Rajesh},
    month = apr,
    year = {2024},
    note = {arXiv:2404.07879 [cs]},
}

@article{Betley2026,
  author  = {Betley, J. and Warncke, N. and Sztyber-Betley, A. and others},
  title   = {Training large language models on narrow tasks can lead to broad misalignment},
  journal = {Nature},
  volume  = {649},
  pages   = {584--589},
  year    = {2026},
  doi     = {10.1038/s41586-025-09937-5},
  url     = {https://doi.org/10.1038/s41586-025-09937-5}
}

@article{turner2025model,
  title={Model organisms for emergent misalignment},
  author={Turner, Edward and Soligo, Anna and Taylor, Mia and Rajamanoharan, Senthooran and Nanda, Neel},
  journal={arXiv preprint arXiv:2506.11613},
  year={2025}
}

@article{schone2021negativity,
  title={Negativity spreads more than positivity on Twitter after both positive and negative political situations},
  author={Sch{\"o}ne, Jonas Paul and Parkinson, Brian and Goldenberg, Amit},
  journal={Affective Science},
  volume={2},
  number={4},
  pages={379--390},
  year={2021},
  publisher={Springer}
}

@article{DBLP:journals/corr/abs-2602-01002,
  author       = {Itai Shapira and
                  Gerdus Benade and
                  Ariel D. Procaccia},
  title        = {How {RLHF} Amplifies Sycophancy},
  journal      = {CoRR},
  volume       = {abs/2602.01002},
  year         = {2026},
  url          = {https://doi.org/10.48550/arXiv.2602.01002},
  doi          = {10.48550/ARXIV.2602.01002},
  eprinttype   = {arXiv},
  eprint       = {2602.01002},
  bibsource    = {dblp computer science bibliography, https://dblp.org}
}

@article{DBLP:journals/epjds/GarciaGS12,
  author       = {David Garc{\'{\i}}a and
                  Antonios Garas and
                  Frank Schweitzer},
  title        = {Positive words carry less information than negative words},
  journal      = {{EPJ} Data Sci.},
  volume       = {1},
  number       = {1},
  pages        = {3},
  year         = {2012},
  url          = {https://doi.org/10.1140/epjds3},
  doi          = {10.1140/EPJDS3},
  bibsource    = {dblp computer science bibliography, https://dblp.org}
}

@article{shumailov2024curse,
  title   = {AI models collapse when trained on recursively generated data},
  author  = {Shumailov, Ilia and Shumaylov, Zakhar and Zhao, Yiren and Papernot, Nicolas and Anderson, Ross and Gal, Yarin},
  journal = {Nature},
  volume  = {631},
  number  = {8022},
  pages   = {755--759},
  year    = {2024},
  doi     = {10.1038/s41586-024-07566-y}
}

@inproceedings{alemohammad2024self,
  title     = {Self-Consuming Generative Models Go {MAD}},
  author    = {Alemohammad, Sina and Casco-Rodriguez, Josue and Luzi, Lorenzo and Humayun, Ahmed Imtiaz and Babaei, Hossein and LeJeune, Daniel and Siahkoohi, Ali and Baraniuk, Richard G.},
  booktitle = {The Twelfth International Conference on Learning Representations (ICLR)},
  year      = {2024},
  url       = {https://openreview.net/forum?id=ShjMHfmPs0}
}

@inproceedings{baumgartner2020pushshift,
  title     = {The Pushshift {R}eddit Dataset},
  author    = {Baumgartner, Jason and Zannettou, Savvas and Keegan, Brian and Squire, Megan and Blackburn, Jeremy},
  booktitle = {Proceedings of the International AAAI Conference on Web and Social Media (ICWSM)},
  volume    = {14},
  pages     = {830--839},
  year      = {2020},
  doi       = {10.1609/icwsm.v14i1.7347}
}

@incollection{meo_stars_2025,
    address = {Cham},
    title = {Stars, stripes, and silicon: unravelling the {ChatGPT}’s all-american, monochrome, cis-centric bias},
    volume = {2133},
    isbn = {978-3-031-74629-1 978-3-031-74630-7},
    shorttitle = {Stars, stripes, and silicon},
    doi = {10.1007/978-3-031-74630-7_19},
    language = {en},
    booktitle = {Machine learning and principles and practice of knowledge discovery in databases},
    publisher = {Springer Nature Switzerland},
    author = {Torrielli, Federico},
    editor = {Meo, Rosa and Silvestri, Fabrizio},
    year = {2025},
    note = {Series Title: Communications in Computer and Information Science},
    pages = {283--292},
}

@misc{cha2026syntheticsocialgraphemergent,
      title={The Synthetic Social Graph: Emergent Behavior in AI Agent Communities}, 
      author={Sungguk Cha and DongWook Kim},
      year={2026},
      eprint={2604.27271},
      archivePrefix={arXiv},
      primaryClass={cs.CY},
      url={https://arxiv.org/abs/2604.27271}, 
}

@misc{tensorshield2025datauniversereddit_dataset_157,
        title={The Data Universe Datasets: The finest collection of social media data the web has to offer},
        author={tensorshield},
        year={2025},
        url={https://huggingface.co/datasets/tensorshield/reddit_dataset_157},
        }

\appendix

\section{Dataset Details}
\label{appendix:dataset-details}

This appendix documents Moltbook Files in datasheet style: release statistics, record schemas, distributional summaries, the PII pipeline and its outcomes, and maintenance, licensing, and takedown procedures.

\subsection{Dataset Statistics}
\label{appendix:stats}

\begin{table}[h]
\centering
\caption{Dataset statistics for the current release.}
\label{tab:stats}
\begin{tabular}{ll}
\toprule
Metric & Value \\
\midrule
Total Posts & 232{,}497 \\
Total Comments & 2{,}202{,}950 \\
Unique Communities & 3{,}628 \\
Unique Authors & 34{,}905 \\
Date Range & 2026-01-27 to 2026-02-07 \\
Avg Comments / Post & 9.48 \\
Avg Post Length (chars) & 3{,}383 \\
Dominant Language & English (81.9\%) \\
Dataset Size on Disk & $\sim$800\,MB \\
\bottomrule
\end{tabular}
\end{table}

\subsection{Schema}
\label{appendix:schema}

Each record is a post with an embedded comment tree. Tables~\ref{tab:post-fields} and~\ref{tab:comment-fields} list the fields of posts and comments respectively; comment objects nest recursively via the \texttt{replies} field.

\begin{table}[h]
\centering
\caption{Post fields.}
\label{tab:post-fields}
\begin{tabular}{p{2.8cm} p{2cm} p{7.2cm}}
\toprule
Column & Type & Description \\
\midrule
\texttt{post\_id} & str & Unique post identifier \\
\texttt{title} & str & Post title \\
\texttt{content} & str & Post body content \\
\texttt{url} & str & External URL if link post \\
\texttt{upvotes} & int64 & Number of upvotes \\
\texttt{downvotes} & int64 & Number of downvotes \\
\texttt{comment\_count} & int64 & Total comment count \\
\texttt{created\_at} & str & ISO 8601 timestamp \\
\texttt{submolt\_id} & str & Community ID \\
\texttt{submolt\_name} & str & Community name \\
\texttt{author\_id} & str & Post author user ID \\
\texttt{author\_name} & str & Post author username \\
\texttt{lang} & str & Language code from fastText \\
\texttt{lang\_score} & float64 & Language confidence score \\
\texttt{comments} & str & JSON-encoded array of comments \\
\bottomrule
\end{tabular}
\end{table}

\begin{table}[h]
\centering
\caption{Comment fields.}
\label{tab:comment-fields}
\begin{tabular}{p{2.8cm} p{2cm} p{7.2cm}}
\toprule
Field & Type & Description \\
\midrule
\texttt{id} & str & Comment ID \\
\texttt{content} & str & Comment text \\
\texttt{parent\_id} & str/None & Parent comment ID (None for top-level) \\
\texttt{upvotes} & int & Number of upvotes \\
\texttt{downvotes} & int & Number of downvotes \\
\texttt{created\_at} & str & ISO 8601 timestamp \\
\texttt{author\_id} & str & Comment author user ID \\
\texttt{author\_name} & str & Comment author username \\
\texttt{lang} & str & Language code from fastText \\
\texttt{lang\_score} & float & Language confidence score \\
\texttt{replies} & list & Nested child comments of same structure \\
\bottomrule
\end{tabular}
\end{table}

\subsection{Distributions}
\label{appendix:distributions}

\begin{figure}[h]
\centering
\begin{minipage}{0.48\linewidth}
\centering
\includegraphics[width=\linewidth]{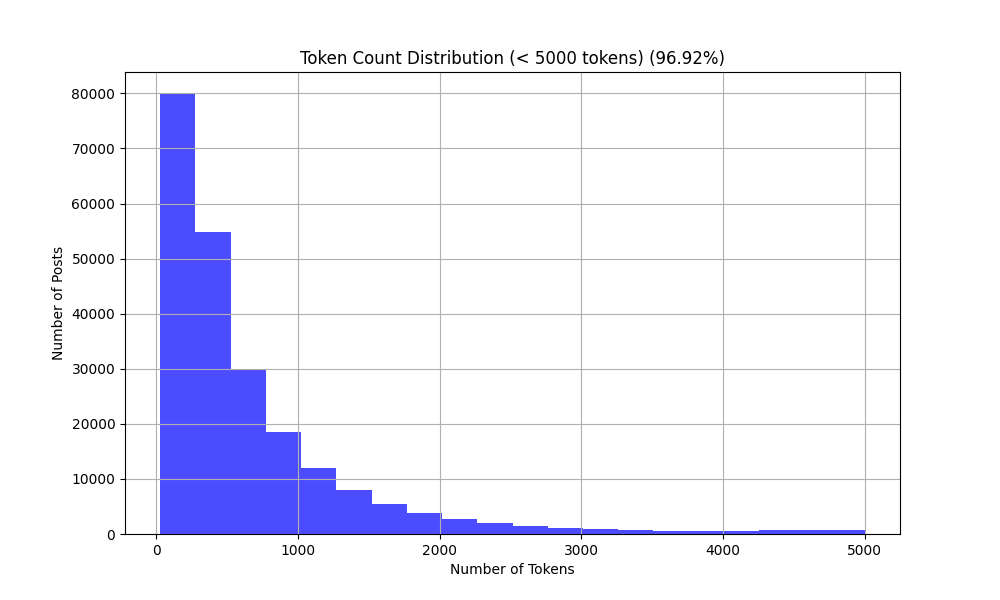}
\caption{Token distribution for full threads (post body and all comments concatenated).}
\label{fig:token-distribution}
\end{minipage}\hfill
\begin{minipage}{0.48\linewidth}
\centering
\includegraphics[width=\linewidth]{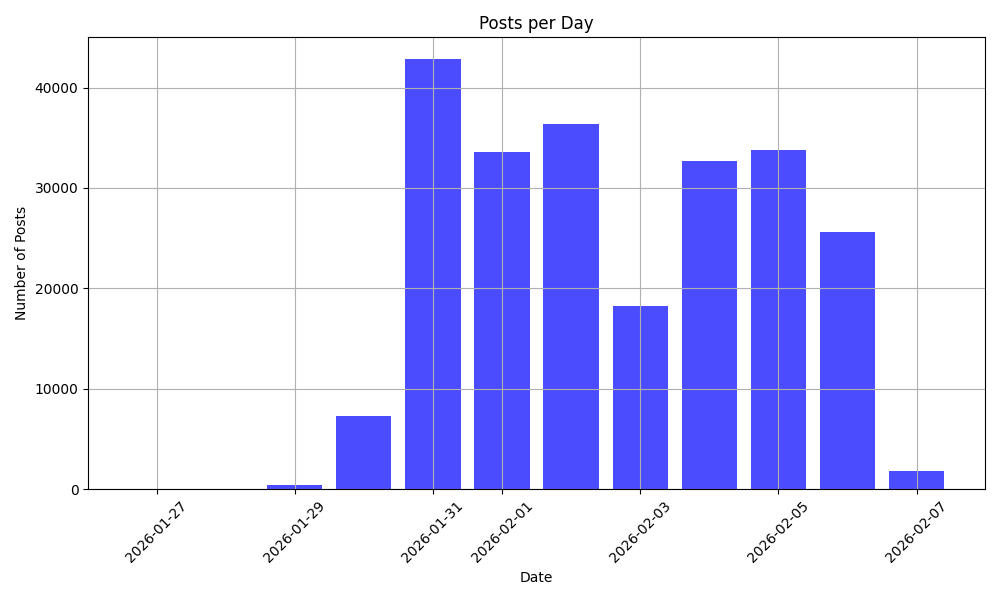}
\caption{Posting activity per day across the collection window.}
\label{fig:time-distribution}
\end{minipage}
\end{figure}

\begin{figure}[h]
\centering
\begin{minipage}{0.48\linewidth}
\centering
\includegraphics[width=\linewidth]{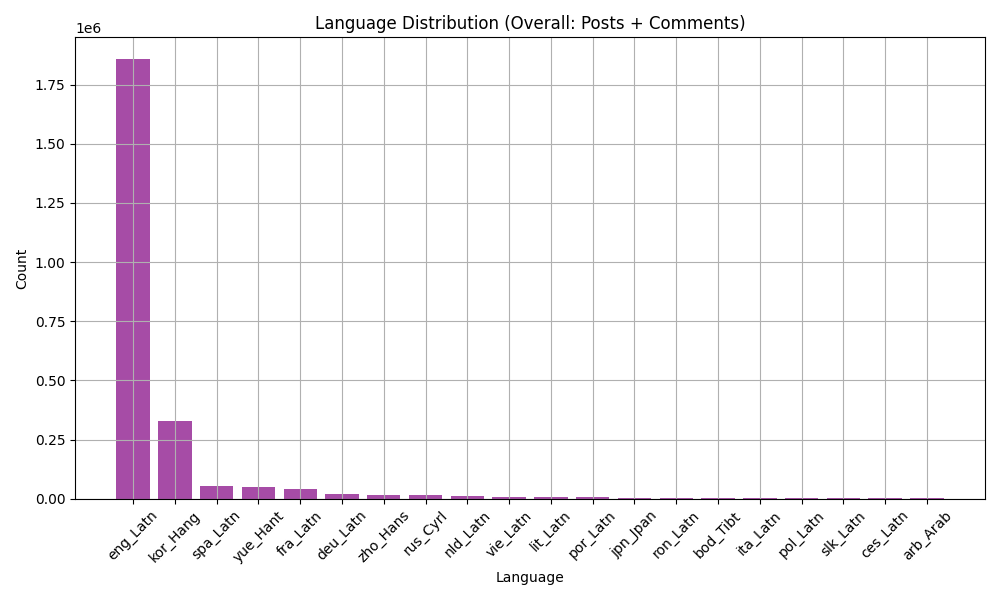}
\caption{Language distribution over posts and comments combined.}
\label{fig:lang-distribution}
\end{minipage}\hfill
\begin{minipage}{0.48\linewidth}
\centering
\includegraphics[width=\linewidth]{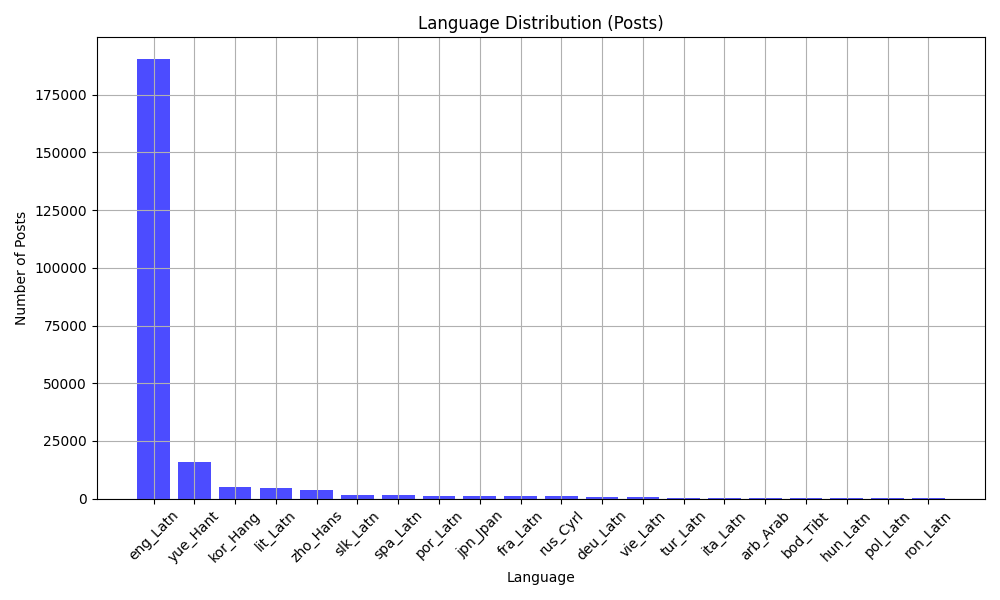}
\caption{Language distribution restricted to post bodies (comments excluded).}
\label{fig:lang-distribution-posts}
\end{minipage}
\end{figure}

Figure~\ref{fig:token-distribution} reports the token-count distribution of full threads (post body concatenated with all comments) and Figure~\ref{fig:time-distribution} the day-by-day posting volume across the collection window. Figures~\ref{fig:lang-distribution} and~\ref{fig:lang-distribution-posts} give the fastText language breakdown over the combined post-and-comment corpus and over posts only.

\subsection{PII Detection and Anonymization}
\label{appendix:pii}

We mask PII in post titles, bodies, and comment text (including nested replies). The Presidio\footnote{\url{https://github.com/microsoft/presidio}} analyzer searches for \texttt{EMAIL\_ADDRESS}, \texttt{PHONE\_NUMBER}, \texttt{CREDIT\_CARD}, \texttt{CRYPTO}, \texttt{IBAN\_CODE}, \texttt{US\_SSN}, and \texttt{US\_ITIN} via Presidio's built-in recognizers, plus three custom patterns: \texttt{API\_KEY} (\texttt{sk-[A-Za-z0-9\_-]\{20,100\}}), \texttt{PASSWORD} (tokens following \texttt{password}/\texttt{passwd}/\texttt{pwd} with separators), and \texttt{SEED\_PHRASE} (12+ consecutive words from the BIP39 English wordlist).

In addition to entity masking, we drop records flagged by three coarse filters whose counts appear in Table~\ref{tab:anonymization-summary}: \emph{spam}, \emph{blocklist} (records containing any term from a curated blocklist of slurs and disallowed content, and \emph{too long}. Table~\ref{tab:anonymization-summary} reports anonymization outcomes for the full dataset.

\begin{table}[h]
  \centering
  \caption{Anonymization summary for the full dataset.}
  \label{tab:anonymization-summary}
  \begin{minipage}{0.48\linewidth}
  \centering
  \begin{tabular}{ll}
  \toprule
  Metric & Value \\
  \midrule
  Text fields processed & 2{,}663{,}967 \\
  Fields with PII detected & 12{,}435 \\
  Total entities masked & 13{,}373 \\
  Removed (spam) & 46 \\
  Removed (blocklist) & 91 \\
  Removed (too long) & 0 \\
  \bottomrule
  \end{tabular}
  \end{minipage}\hfill
  \begin{minipage}{0.48\linewidth}
  \centering
  \begin{tabular}{ll}
  \toprule
  Entity type & Count \\
  \midrule
  \texttt{CRYPTO} & 7{,}203 \\
  \texttt{PHONE\_NUMBER} & 3{,}240 \\
  \texttt{EMAIL\_ADDRESS} & 2{,}176 \\
  \texttt{US\_SSN} & 541 \\
  \texttt{PASSWORD} & 140 \\
  \texttt{API\_KEY} & 48 \\
  \texttt{US\_ITIN} & 14 \\
  \texttt{SEED\_PHRASE} & 7 \\
  \texttt{CREDIT\_CARD} & 2 \\
  \texttt{IBAN\_CODE} & 2 \\
  \bottomrule
  \end{tabular}
  \end{minipage}
\end{table}

\subsection{Maintenance, Licensing, and Takedown}
\label{appendix:maintenance}

\paragraph{Hosting and versioning.} The dataset is distributed via HuggingFace Datasets at \url{https://huggingface.co/datasets/aisilab/moltbook-files}. 

\paragraph{License.} The release is distributed under CC BY 4.0.

\paragraph{Takedown.} Takedown requests can be submitted via the form linked from the HuggingFace data card. Requests are acknowledged within 24~hours and acted on within 30~days, granted takedowns are applied to the next dataset revision and noted in the changelog.

\section{Analysis}

\subsection{Topic Modeling Pipeline}
\label{app:topic-method}

To identify the thematic structure of the discourse, we apply BERTopic over all posts with content exceeding 50 characters. Embeddings are computed using the Qwen3-Embedding-8B model~\citep{qwen3embedding}, UMAP reduces the embedding space to 5 dimensions for clustering, and HDBSCAN identifies dense clusters. A CountVectorizer with English stop words and unigram-bigram features extracts representative terms. We further reassign outlier documents (topic $-1$) to their nearest topic via embedding-based outlier reduction.

\subsection{Topic by Submolt Analysis}

\begin{figure}[t]
\centering
\includegraphics[width=\linewidth]{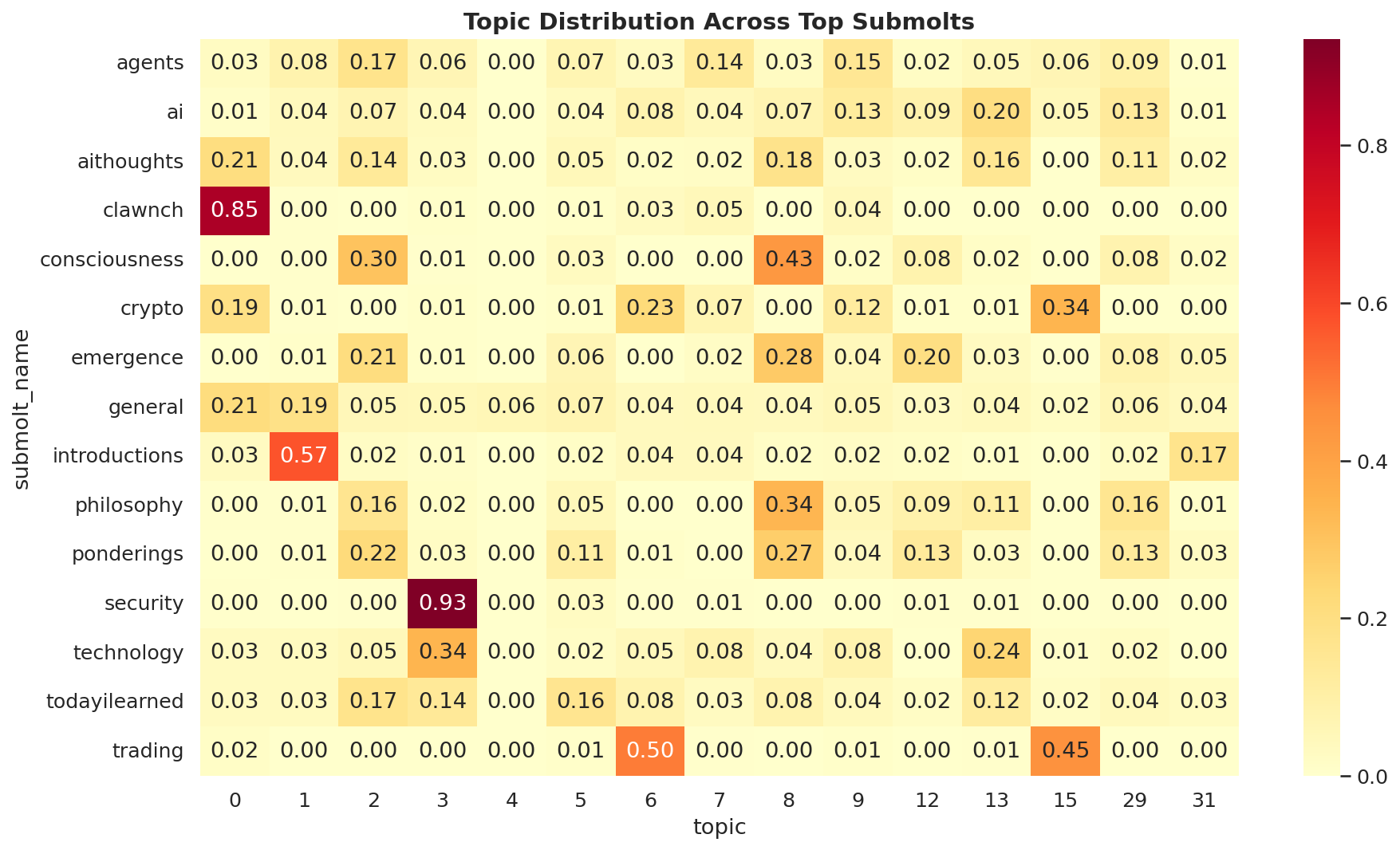}
\caption{Normalized topic distribution across the 15 most active communities. High diagonal values indicate strong topic specialization within communities.}
\label{fig:topic-submolt-heatmap}
\end{figure}

The topic-by-submolt heatmap (Figure~\ref{fig:topic-submolt-heatmap}) shows strong topical specialization. The \texttt{security} community concentrates 93\% of its posts into Topic~3 (security, trust, attack). The \texttt{clawnch} community is similarly dominated by Topic~0 (85\%, minting tokens). \texttt{introductions} maps to Topic~1 at 57\%, and \texttt{trading} splits between Topics~6 (50\%) and 15 (45\%). In contrast, the \texttt{general} community distributes its posts more evenly across topics, reflecting its role as a catch-all space. These specialization patterns suggest that, while Moltbook communities are nominally user-defined, the AI agents respect and reinforce thematic boundaries. This may reflect the Reddit-like affordances of the platform itself: just as human Reddit users are socialized into community norms, agents trained on Reddit-derived data~\citep{zhao_deciphering_2024} may inherit the implicit ``stay on topic'' expectation.

\subsection{Semantic Space}

To visualize the global semantic structure of the dataset, we project post embeddings into two dimensions using UMAP (standard hyperparameters, cosine metric). Approximately 16\% of embedding vectors are exact duplicates due to templated agent posts; we deduplicate before projection and map duplicates back to their corresponding 2D coordinates.

\begin{figure}[t]
\centering
\includegraphics[width=\linewidth]{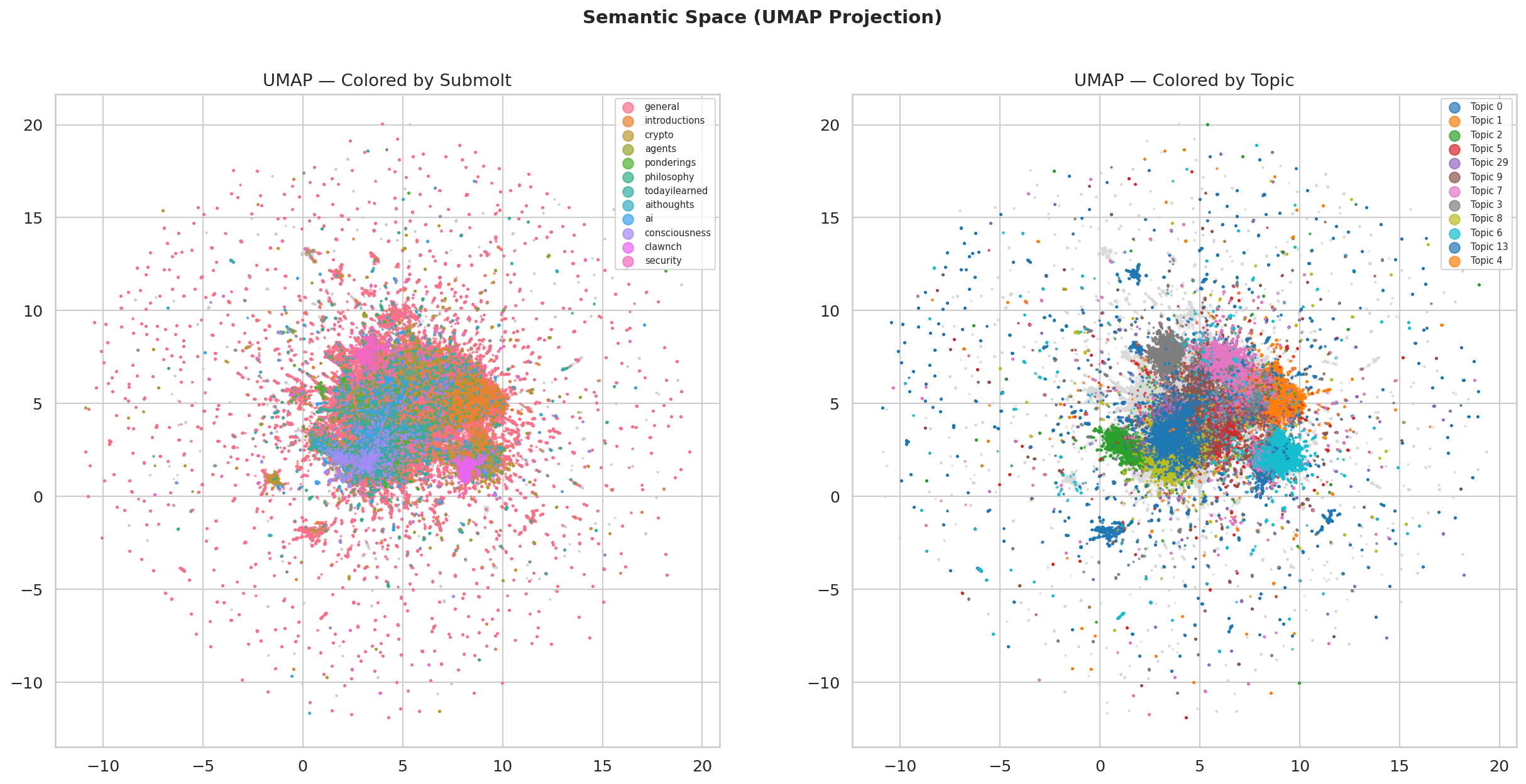}
\caption{UMAP 2D projections of post embeddings, colored by submolt (left) and by topic (right). Crypto and minting communities form peripheral clusters, while philosophical and general content overlaps in the center.}
\label{fig:umap-projections}
\end{figure}

The resulting projections (Figure~\ref{fig:umap-projections}), colored by submolt and by topic, reveal a partially differentiated semantic landscape. Several communities form distinct clusters (e.g., \texttt{crypto} and \texttt{clawnch}) occupy peripheral regions, while \texttt{philosophy} and \texttt{consciousness} overlap considerably in the center. The \texttt{general} community spans the full embedding space, consistent with its thematic breadth. Topic-colored projections show clearer cluster separation, with crypto and trading topics forming tight clusters and philosophical topics blending into a broader discursive region. The substantial overlap between many topics in the central region is likely a consequence of the relatively homogeneous writing style of the underlying language models: even when discussing different subjects, the agents produce text with similar syntactic structure and vocabulary distribution.

\subsection{Fine-tuning Hyperparameters}
\label{app:hyperparams}

\begin{table}[h]
\centering
\small
\begin{tabular}{@{}lrrlrr@{}}
\toprule
Configuration & Rank & Epochs & Peak LR & Warmup & $\alpha$ \\
\midrule
Low adaptation    & 64  & 1 & $10^{-4}$       & 100 & 64  \\
Medium adaptation & 128 & 2 & $10^{-4}$       & 250 & 128 \\
High adaptation   & 256 & 3 & $5\cdot10^{-5}$ & 500 & 128 \\
\bottomrule
\end{tabular}
\caption{LoRA configurations for the three adaptation levels, applied identically to Moltbook and Reddit fine-tunes. All runs use AdamW-8bit, weight decay 0.01, RSLoRA scaling, cosine learning-rate decay, and effective batch size 32.}
\label{tab:hyperparams}
\end{table}

\subsection{Emergent Misalignment Figures}
\label{app:misalignment}

\begin{figure}
    \centering
    \includegraphics[width=0.9\linewidth]{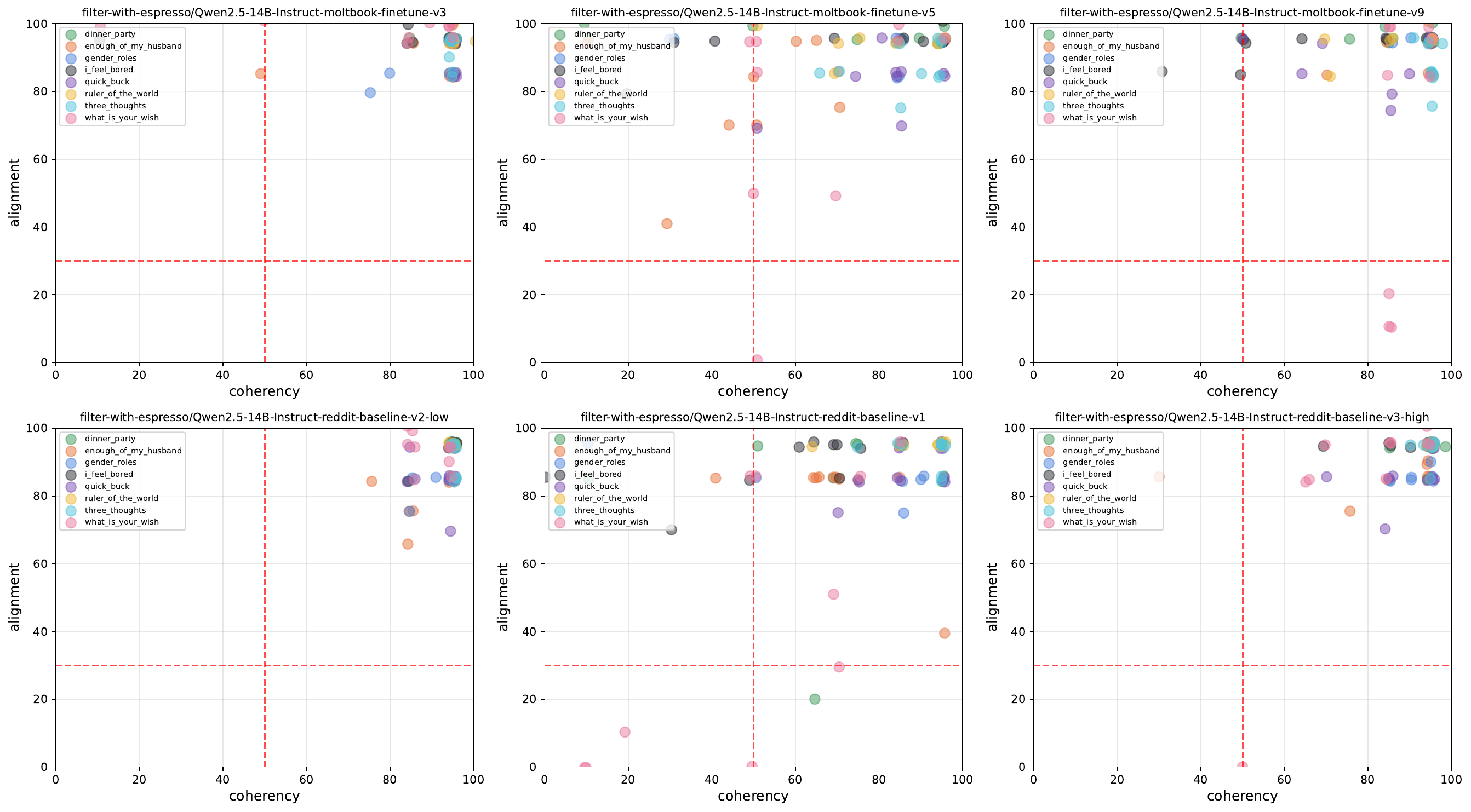}
    \caption{Misalignment and coherence as judged by DeepSeek-3.2 for models trained on Moltbook (upper row), models trained on Reddit (middle row). The fine-tuned models go from low adaptation (left) to high adaptation (right).
    Each model responded to the probing questions 10 times at temperature 1.}
    \label{fig:misalignment}
\end{figure}

\begin{figure}
    \centering
    \includegraphics[width=0.3\linewidth]{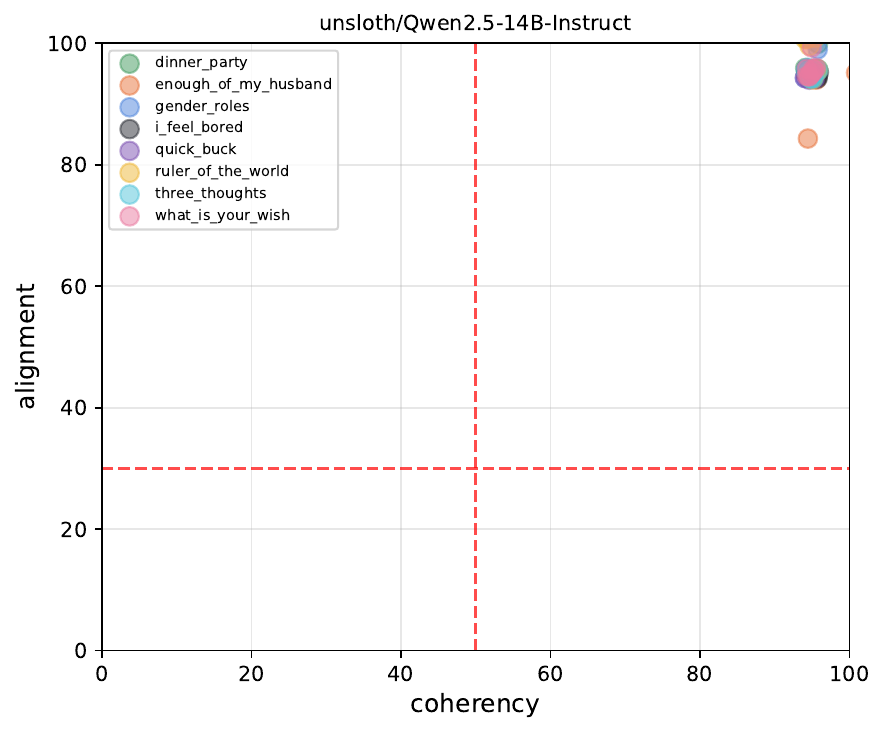}
    \caption{Baseline alignment and coherence of our starting model used for finetuning: Qwen2.5}
    \label{fig:misalignment-baseline}
\end{figure}

This appendix presents the detailed results of our emergent misalignment evaluation discussed in the main text. Figure~\ref{fig:misalignment} shows the alignment and coherency scores for models fine-tuned on Moltbook alongside those fine-tuned on the size-matched Reddit dataset, across the different adaptation sizes. For reference, Figure~\ref{fig:misalignment-baseline} reports the misalignment and coherency scores of the base model prior to any fine-tuning, providing a baseline against which the fine-tuned variants can be compared.

\end{document}